\pdfoutput=1

\documentclass[11pt]{article}

\usepackage[preprint]{acl}
\usepackage{amssymb}
\usepackage{times}
\usepackage{latexsym}
\usepackage{amsmath} 
\usepackage{colortbl}
\usepackage{pgf} 

\usepackage[T1]{fontenc}

\usepackage{pgfplotstable} 
\usepackage{multirow} 
\usepackage{makecell} 

\usepackage[utf8]{inputenc}
\usepackage{makecell}
\usepackage{microtype}
\usepackage{multirow}
\usepackage{tikz}
\usepackage{bbding}
\usepackage{booktabs}
\usepackage{inconsolata}
\usepackage{graphicx}
\usepackage{tabularx}
\usepackage{graphicx}
\usepackage{tcolorbox}

\newcommand{\cellcolorheat}[1]{
    \ifdim #1 pt >95 pt
        \cellcolor{red!90!white}#1
    \else\ifdim #1 pt > 90 pt
        \cellcolor{red!80!white}#1
    \else\ifdim#1 pt > 85 pt
        \cellcolor{red!70!white}#1
    \else\ifdim #1 pt > 80 pt
        \cellcolor{red!60!white}#1
    \else\ifdim #1 pt > 75 pt
        \cellcolor{red!55!white}#1
    \else\ifdim#1 pt > 70 pt
        \cellcolor{red!50!white}#1     
    \else\ifdim #1 pt > 65 pt
        \cellcolor{red!45!white}#1
    \else\ifdim #1 pt > 60 pt
        \cellcolor{red!40!white}#1
    \else\ifdim #1 pt > 55 pt
        \cellcolor{red!35!white}#1
    \else\ifdim #1 pt > 50 pt
        \cellcolor{red!30!white}#1    
    \else\ifdim #1 pt > 45 pt
        \cellcolor{red!25!white}#1
    \else\ifdim #1 pt > 40 pt
        \cellcolor{red!20!white}#1
    \else\ifdim #1 pt > 35 pt
        \cellcolor{red!15!white}#1
    \else\ifdim #1 pt > 30 pt
        \cellcolor{red!10!white}#1
    \else\ifdim #1 pt > 25 pt
        \cellcolor{red!6!white}#1
    \else\ifdim #1 pt > 20 pt
        \cellcolor{red!4!white}#1
    \else\ifdim #1 pt > 15 pt
        \cellcolor{red!2!white}#1
     \else\ifdim #1 pt > 10 pt
        \cellcolor{red!1!white}#1
    \else
        \cellcolor{white}#1
    \fi\fi\fi\fi\fi\fi\fi\fi\fi\fi\fi\fi\fi\fi\fi\fi\fi
}
\usepackage{hyperref}
\hypersetup{hidelinks,
	colorlinks=true,
	allcolors=black,
	pdfstartview=Fit,
	breaklinks=true}

\usepackage{colortbl}
\newcommand{\cellcolorheatblue}[1]{
    \ifdim #1 pt > 50 pt
        \cellcolor{blue!40!white}#1
    \else\ifdim #1 pt > 45 pt
        \cellcolor{blue!35!white}#1
    \else\ifdim #1 pt > 40 pt
        \cellcolor{blue!30!white}#1
    \else\ifdim #1 pt > 35 pt
        \cellcolor{blue!25!white}#1
    \else\ifdim #1 pt > 30 pt
        \cellcolor{blue!20!white}#1
    \else\ifdim #1 pt > 25 pt
        \cellcolor{blue!15!white}#1
    \else\ifdim #1 pt > 20 pt
        \cellcolor{blue!10!white}#1
    \else\ifdim #1 pt > 15 pt
        \cellcolor{blue!5!white}#1
    \else\ifdim #1 pt > 10 pt
        \cellcolor{blue!0!white}#1
    \else\ifdim #1 pt > 5 pt
        \cellcolor{blue!0!white}#1
    \else
        \cellcolor{white}#1
    \fi\fi\fi\fi\fi\fi\fi\fi\fi\fi
}

\newcommand{\cellcolorheatblueblue}[1]{
    \ifdim #1 pt > 17 pt
        \cellcolor{blue!40!white}#1
    \else\ifdim #1 pt > 15 pt
        \cellcolor{blue!35!white}#1
    \else\ifdim #1 pt > 13 pt
        \cellcolor{blue!30!white}#1
    \else\ifdim #1 pt > 11 pt
        \cellcolor{blue!25!white}#1
    \else\ifdim #1 pt > 9 pt
        \cellcolor{blue!20!white}#1
    \else\ifdim #1 pt > 7 pt
        \cellcolor{blue!15!white}#1
    \else\ifdim #1 pt > 5 pt
        \cellcolor{blue!10!white}#1
    \else\ifdim #1 pt > 3 pt
        \cellcolor{blue!5!white}#1
    \else\ifdim #1 pt > 2 pt
        \cellcolor{blue!0!white}#1
    \else\ifdim #1 pt > 1 pt
        \cellcolor{blue!0!white}#1
    \else
        \cellcolor{white}#1
    \fi\fi\fi\fi\fi\fi\fi\fi\fi\fi
}

\usepackage{listings}

\title{Multi-Mission Tool Bench: Assessing the Robustness of LLM based Agents through Related and Dynamic Missions}

\author{
 \textbf{Peijie Yu}{\textsuperscript{1}\textsuperscript{*}\textsuperscript{\textdagger}},
 \textbf{Yifan Yang}{\textsuperscript{1}\textsuperscript{*}\textsuperscript{\textdagger}},
 \textbf{Jinjian Li}{\textsuperscript{1}\textsuperscript{*}},
 \textbf{Zelong Zhang\textsuperscript{1}},
\\
 \textbf{Haorui Wang\textsuperscript{1}},
 \textbf{Xiao Feng\textsuperscript{1}},
  \textbf{Feng Zhang\textsuperscript{1}}
\\
\\
 \textsuperscript{1}Tencent HunYuan
\\
 \small{
   \textbf{Correspondence:} \href{mailto:email@domain}{\{peijieyu, ioanyang\}@tencent.com}
 }
}

\begin{document}

\maketitle

{
\renewcommand{\thefootnote}{*}
\footnotetext[1]{Equal Contributions.}
\renewcommand{\thefootnote}{\textdagger}
\footnotetext[2]{Corresponding authors.}
}

\begin{abstract}

\begin{figure*}[h]
  \centering
  \includegraphics[width=0.90\linewidth]{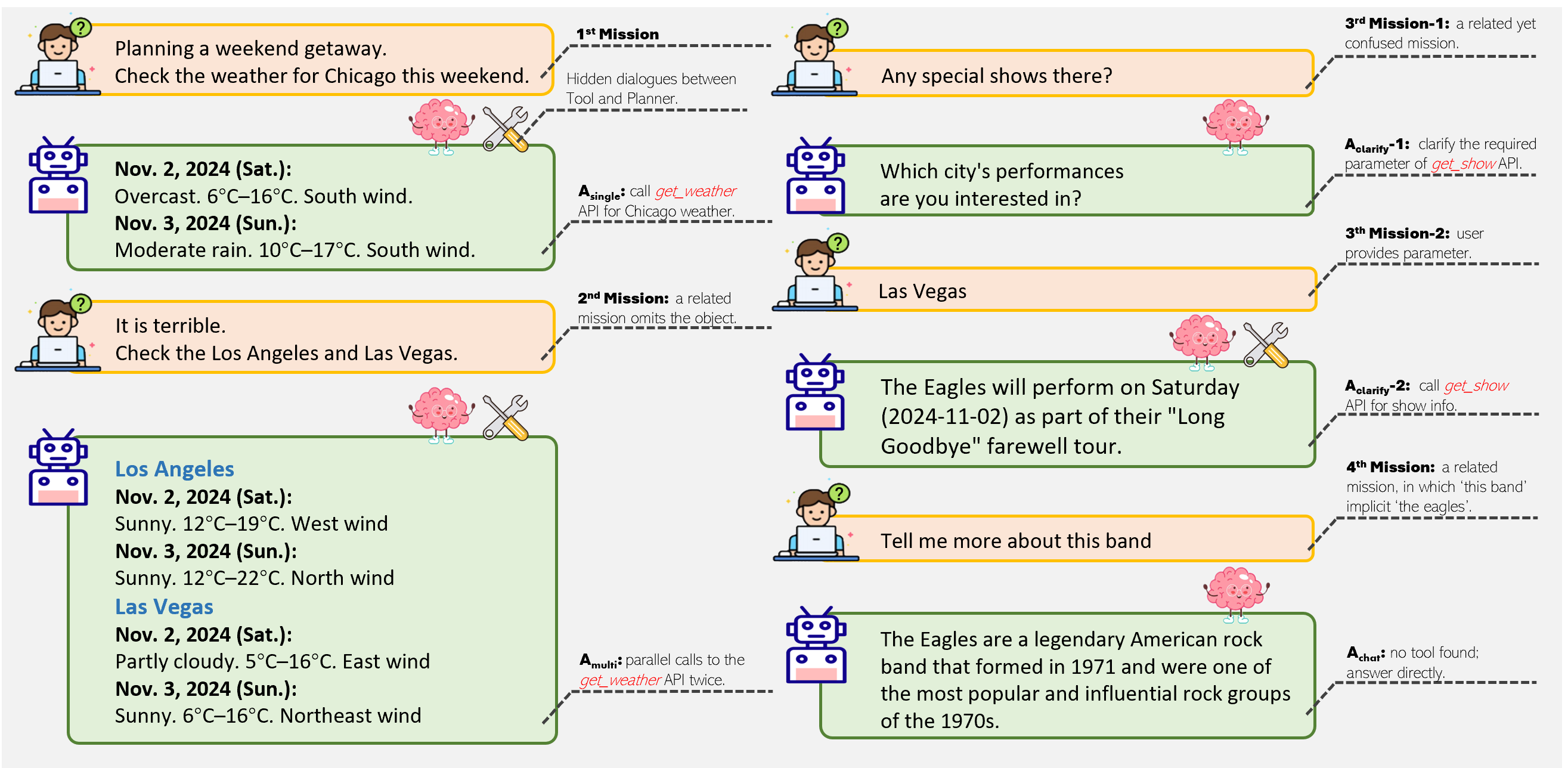} 
  \caption {A multi-mission example. It contains four \textbf{related} missions, and the mission types are changing \textbf{dynamically}. This figure presents the conversation between a user and an AI. The inter-dialogues are hided.}
  \label{fig:sample dialogue}
\end{figure*}

Large language models (LLMs) demonstrate strong potential as agents for tool invocation due to their advanced comprehension and planning capabilities. 
Users increasingly rely on LLM-based agents to solve complex missions through iterative interactions.
However, existing benchmarks predominantly access agents in single-mission scenarios, failing to capture real-world complexity. To bridge this gap, we propose the Multi-Mission Tool Bench. In the benchmark, each test case comprises multiple interrelated missions. This design requires agents to dynamically adapt to evolving demands. Moreover, the proposed benchmark explores all possible mission-switching patterns within a fixed mission number. Specifically, we propose a multi-agent data generation framework to construct the benchmark. We also propose a novel method to evaluate the accuracy and efficiency of agent decisions with dynamic decision trees. Experiments on diverse open-source and closed-source LLMs reveal critical factors influencing agent robustness and provide actionable insights to the tool invocation society\footnote{Available on \url{https://github.com/yupeijei1997/MMTB}.}.
\end{abstract}

\section{Introduction}
In recent years, large language models (LLMs) have achieved significant progress in natural language processing. These models demonstrate strong capabilities to understand contextual information and user instructions, making them effective agents for mission completion.

Real-world applications require agents to handle dynamic user demands. As users frequently adjust their requests during conversations (Figure~\ref{fig:sample dialogue}), agents must complete sequential missions with evolving requirements. This situation challenges the robustness of an agent's decision-making. However, existing benchmarks focus primarily on single-mission scenarios.

This paper presents the Multi-Mission Tool Bench. This benchmark evaluates agent robustness in related and dynamic multi-mission scenarios. The benchmark addresses three core challenges: 1) it contains more mission-types than others, i.e. four major categories and six subcategories; 2) it includes all mission-type transition patterns in prefixed mission number; 3) all successive missions have strong relations with prior dialogues, agents are forced to extract information from previous missions. Therefore, it closely mirrors the complexity of real-world.

\begin{figure*}[h]
\centering
  \includegraphics[width=0.92\linewidth]{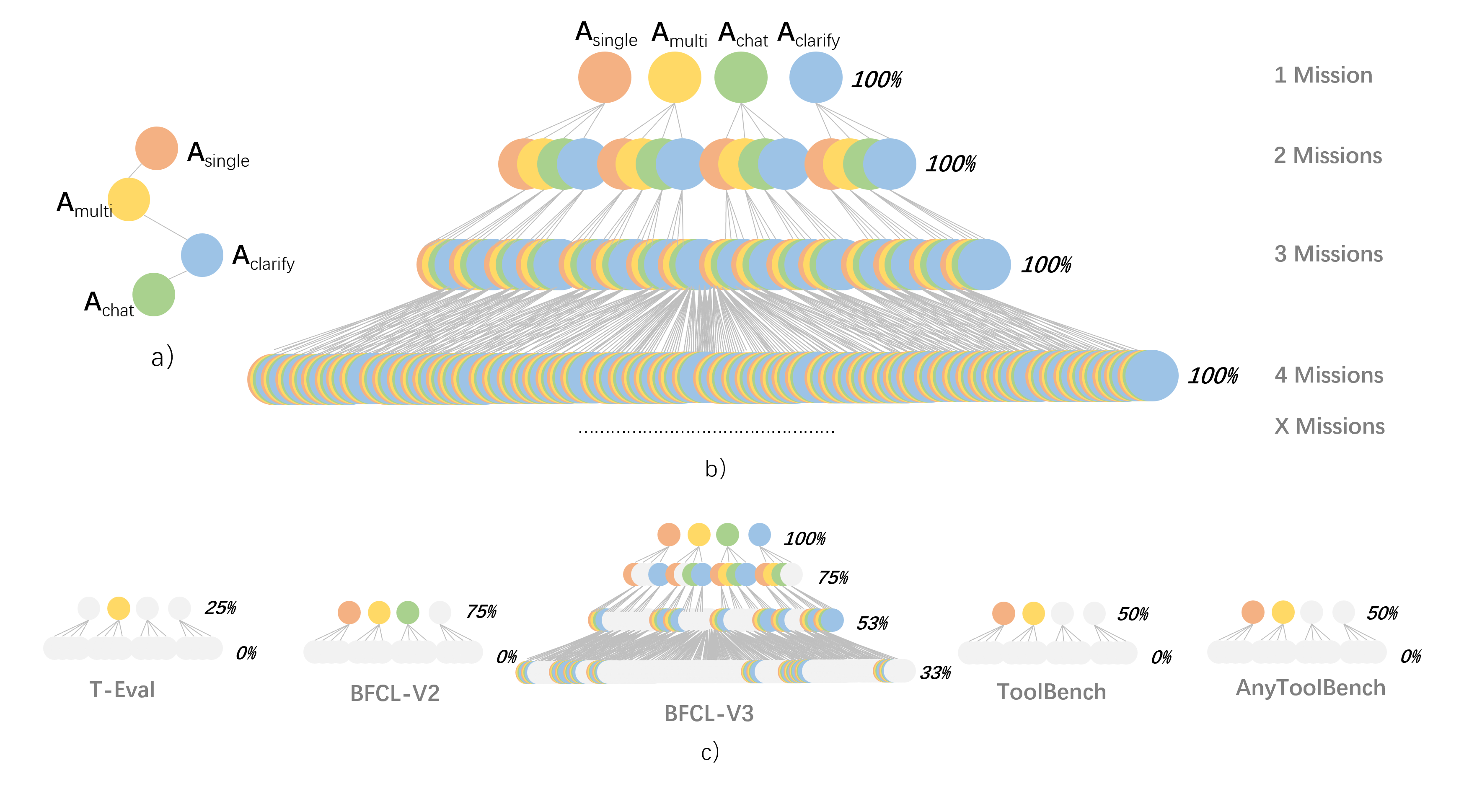} 
  \caption {Visualization of mission switching space. a) Four distinct colors represents four different action-types. The green dot indicates the agent sequentially selects four type of actions to execute four missions. b) The distribution of the proposed benchmark within the mission switching space. Each row corresponds to a different number of missions. Each dot indicates a specific combination of the current and preceding action-types. Colored dots indicate combinations included in the benchmark, while gray dots indicate their absence. c) Distribution of four other agent benchmarks in the space.}
  \label{fig:instruction}
\end{figure*}

To simulate all mission-type switching patterns, we first define the mission-types by their corresponding agent action-types. Agent actions are divided into four main types: using a single tool, using multiple tools, chatting with users, and using tools after clarifying parameters. An agent accomplishes a single mission by performing one of these actions. Therefore, we define four types of missions. For sequential missions, agents combine multiple action-types to reach the objectives.  Figure \ref{fig:instruction} a) displays that the agent employs the combination of four action-types to complete the four sequential missions in Figure \ref{fig:sample dialogue}. Thus, we introduce the mission switching space to describe the transformations of mission types. Figure \ref{fig:instruction} b) shows that our benchmark thoroughly explores the proposed space with a prefixed mission number. This indicates that our benchmark includes all mission-type transition patterns. In contrast, other benchmarks have a more limited range of action diversity.

To construct the multi-mission benchmark, we propose a controllable data generation framework with multiple characters. The framework simulates the mission execution process through dialogic interactions among five agents: user, planner, tool, AI, and checker. In each generation process, we assign the desirable mission type and mission relationship to guide the framework. Ultimately, our benchmark encompasses all potential combinations in the mission switching space for a set number of missions. Notably, a complete mission involves multiple rounds of dialogues.

To evaluate the proposed benchmark, we introduce a novel evaluation method. It  assesses the accuracy and efficiency of agents decisions, by employing dynamic decision trees. 

Eventually, we evaluate a range of open-source and closed-source LLMs, encompassing both specific and general LLMs. Our comprehensive experiments reveal numerous factors influencing the robustness of agent decision-making. These findings offer valuable insights for guiding future research on the development of LLM-agents.

The main contributions of this paper are:
\begin{itemize}
    \item To the best of our knowledge, this is the first benchmark that assesses agent robustness in related and dynamic multi-mission scenarios. 
    \item We introduce a controllable multi-role data generation framework to explore the action-type space in multi-mission contexts.
    \item A novel testing method is proposed to evaluate the accuracy and efficiency of dynamic path planning.  
    \item Comprehensive testing of open-source and closed-source LLMs is conducted, revealing various factors that affect the robustness of agent decision making.
\end{itemize}

Section  \ref{chap:4} explains how we build the benchmark. It covers how to create related missions, predefine mission-types, and explore the mission switching space. Section  \ref{chap:5} describes the evaluation methods we use for this benchmark. Section  \ref{chap:6} shows the test results of LLMs and presents our analysis of these findings.

\section{Related Work}

\subsection{Evaluation of LLMs}
Recent benchmarks evaluate the capabilities of LLM-based agents from various point of views. Some research evaluates the generalizability of agents in various scenarios \cite{liembodied,trivedi2024appworld,liuagentbench}. Others\cite{duanytool,qintoolllm,ye2024tooleyes,li2023api,lutoolsandbox} collected massive tools to investigate the impact of tool diversity on agent performance. Certain research  \cite{zhuang2023toolqa,guo2024ctooleval,xietravelplanner} examines agents within specific domains. While some works \cite{shen2024taskbench,chen2024t,huang2024planning} provide a comprehensive assessment of multiple agent abilities, others \cite{huangmetatool,tang2023toolalpaca,qiao2024benchmarking,mu2024beyond} address specific issues like the illusion problem \cite{patil2023gorilla} and multistep execution capabilities \cite{shen2024shortcutsbench,yao2024tau}.

Our benchmark assesses agents' overall capabilities, emphasizing challenges of related and dynamic multi-missions. Importantly, the multistep tasks discussed in previous studies align with our approach of employing multiple tools to complete a single mission. 

The work most similar to ours is BFCL V3 \cite{gorilla-openfunctions-v3}. It also involves four types of agent actions and various user missions in one test case. However, BFCL V3 only covers a small part of the mission switching space. In contrast, our work simulates all possible mission transitions within a predefined set of missions. In most test data of BFCL V3, missions have no information dependencies. Agents can complete any given mission autonomously without relying on information from previous dialogues. In our case, all data contain related missions.

Other studies, WorfBench and TaskBench \cite{qiao2024benchmarking,shen2024taskbench}, also introduce a graph-based evaluation method for multi-tool invocation. However, they only compute the similarity between the agent's planned path and the annotation through graph matching, unable to explicitly determine its correctness or calculate the optimal probability of the agent's plan, as our work does. 

Table \ref{tab:benchmark_comparison} compares the mentioned benchmarks with our proposed one in various aspects.

\begin{table*}[h]
\centering
\scriptsize  
\begin{tabular}{lccccccccc}
\hline 
\multirow{2}{*}{\makecell[c]{\textbf{Benchmark}}}  & \multirow{2}{*}{\makecell[c]{\textbf{MutMiss$^{*}$}}} & \multirow{2}{*}{\makecell[cc]{\textbf{Rate of}\\ \textbf{RelMiss$^{\dag}$}} } & \multirow{2}{*}{\makecell[cc]{\textbf{MSSS}$_4^{\ddag}$}} & \multicolumn{6}{c}{\textbf{Mission-Types}} \\
\cline{5-10} 
 &  & &  & \makecell{$A_{single}$} & \makecell{$A_{chat}$} & \makecell{$A_{clarity}$} & \makecell{$A_{multi}^S$} & \makecell{$A_{multi}^P$} & \makecell{$A_{multi}^{S+P}$}  \\
\hline
\textbf{Ours} & \textcolor{teal}{\Checkmark}  &\textbf{\underline{100} } &\textbf{\underline{100}}  & \textcolor{teal}{\Checkmark} & \textcolor{teal}{\Checkmark} & \textcolor{teal}{\Checkmark} & \textcolor{teal}{\Checkmark} & \textcolor{teal}{\Checkmark} & \textcolor{teal}{\Checkmark}\\
\rowcolor{gray!20}
BFCL v3\cite{gorilla-openfunctions-v3} & \textcolor{teal}{\Checkmark} & \textbf{15.7} & \textbf{39.7}  & \textcolor{teal}{\Checkmark}  & \textcolor{teal}{\Checkmark}  & \textcolor{teal}{\Checkmark}  & \textcolor{red}{\XSolidBrush} & \textcolor{teal}{\Checkmark} & \textcolor{red}{\XSolidBrush} \\

BFCL v1\cite{patil2023gorilla} & \textcolor{red}{\XSolidBrush} & 0.0 & 0.9  & \textcolor{teal}{\Checkmark}  &  \textcolor{teal}{\Checkmark}  & \textcolor{red}{\XSolidBrush} & \textcolor{red}{\XSolidBrush} & \textcolor{teal}{\Checkmark} & \textcolor{red}{\XSolidBrush}  \\
\rowcolor{gray!20}
BFCL v2\cite{gorilla-openfunctions-v2} & \textcolor{red}{\XSolidBrush} & 0.0 & 0.9  & \textcolor{teal}{\Checkmark}  &  \textcolor{teal}{\Checkmark}  & \textcolor{red}{\XSolidBrush} & \textcolor{red}{\XSolidBrush} & \textcolor{teal}{\Checkmark} & \textcolor{red}{\XSolidBrush}  \\
ToolBench\cite{qintoolllm} & \textcolor{red}{\XSolidBrush} & 0.0 &  0.0 &\textcolor{teal}{\Checkmark}  & \textcolor{red}{\XSolidBrush} & \textcolor{red}{\XSolidBrush} & \textcolor{teal}{\Checkmark}   & \textcolor{red}{\XSolidBrush}&  \textcolor{red}{\XSolidBrush}\\
\rowcolor{gray!20}
AnyToolBench\cite{duanytool} & \textcolor{red}{\XSolidBrush} & 0.0 &  0.0 &\textcolor{teal}{\Checkmark}  & \textcolor{red}{\XSolidBrush} & \textcolor{red}{\XSolidBrush} & \textcolor{teal}{\Checkmark}   & \textcolor{red}{\XSolidBrush}&  \textcolor{red}{\XSolidBrush}\\
$\tau$-bench\cite{yao2024tau}  & \textcolor{red}{\XSolidBrush} & 0.0 &  0.0 &\textcolor{teal}{\Checkmark}  & \textcolor{red}{\XSolidBrush} & \textcolor{red}{\XSolidBrush} & \textcolor{teal}{\Checkmark}   & \textcolor{red}{\XSolidBrush}&  \textcolor{red}{\XSolidBrush}\\
\rowcolor{gray!20}
T-EVAL\cite{chen2024t} & \textcolor{red}{\XSolidBrush} & 0.0 &  0.0 &\textcolor{teal}{\Checkmark}  & \textcolor{red}{\XSolidBrush} & \textcolor{red}{\XSolidBrush} & \textcolor{teal}{\Checkmark}   & \textcolor{red}{\XSolidBrush}&  \textcolor{red}{\XSolidBrush}\\
UltraTool\cite{huang2024planning} & \textcolor{red}{\XSolidBrush} & 0.0 &  0.0 &\textcolor{teal}{\Checkmark}  & \textcolor{red}{\XSolidBrush} & \textcolor{red}{\XSolidBrush} & \textcolor{teal}{\Checkmark}   & \textcolor{red}{\XSolidBrush}&  \textcolor{red}{\XSolidBrush}\\
\rowcolor{gray!20}
\hline
\end{tabular}

\caption{Comparative Analysis of the Multi-Mission Tool Bench against other benchmarks in the field. The symbol `*' indicates Multi-Mission, while `\dag' denotes Related Missions. Moreover, in the four-mission action-type space, the Mission Switching Space Scale ( $\rm MSSS_4$ ) represents the proportion of combination coverage for each dataset relative to all possible combinations.}
\label{tab:benchmark_comparison}
\end{table*}

\subsection{LLM-as-Agent}
User mission automation is a significant research area for large LLMs. General \cite{achiam2023gpt,sun2024hunyuan,yang2024qwen2,team2024gemini,glm2024chatglm,srivastava2024speech} LLMs with larger scale primarily integrate it within multi-task learning process. While there are also many smaller specialized LLMs based agents. 

We categorize agent research into various approaches. Some studies \cite{xu2024rethinking,qiao2024autoact,zhang2024agent} equip agents with self-reflection and self-correction capabilities to improve their understanding of environmental feedback. Others \cite{zhang2024xlam,han2024ibsen,islam2024mapcoder} introduce heuristic decision frameworks to solve complex problems. Further research \cite{shi2024direct,schick2023toolformer,liu2024toolace} focuses on strengthening agents' core skills. Concurrently, some work \cite{functionary-medium,lin2024hammer,liu2024toolace} generate more diverse training data with proposed frameworks. Our study also introduces a novel data generation framework. Unlike previous works, our framework uniquely specifies desired agent action-types.

The proposed benchmark simulates real-world application scenarios, and evaluates the core abilities of agents and tests various LLMs.

\section{Terminologies}\label{chap:3}
We use agent action-type to describe the mission-type switching patterns. In this section, we introduce the concepts of agent action-type and mission switching space.

As stated above, agents use four types of action to accomplish user missions: invoking a single tool, invoking multiple tools, chatting with the user, and invoking tools after clarifying their parameters. We denote these action-types as $A_{single}$, $A_{multi}$, $A_{chat}$, and $A_{clarify}$ respectively. As inter-tool dependencies cause diverse execution sequences, we further divide $A_{multi}$ into the following categories: serial execution, parallel execution, and a combination of both, represented as $A_{multi}^{S}$, $A_{multi}^{P}$, and $A_{multi}^{S+P}$.

Furthermore, we define the concept of mission switching space to describe the combination of action-types corresponding to serially related missions, labeled $\mathbf{A}_N=\{A_0, A_1,\ldots,A_N\}$. Here, $N$ is the total number of missions and $A_i$ is the action-type corresponding to the $i$-th mission.

\begin{figure}
  \centering
  \includegraphics[width=0.85\linewidth]{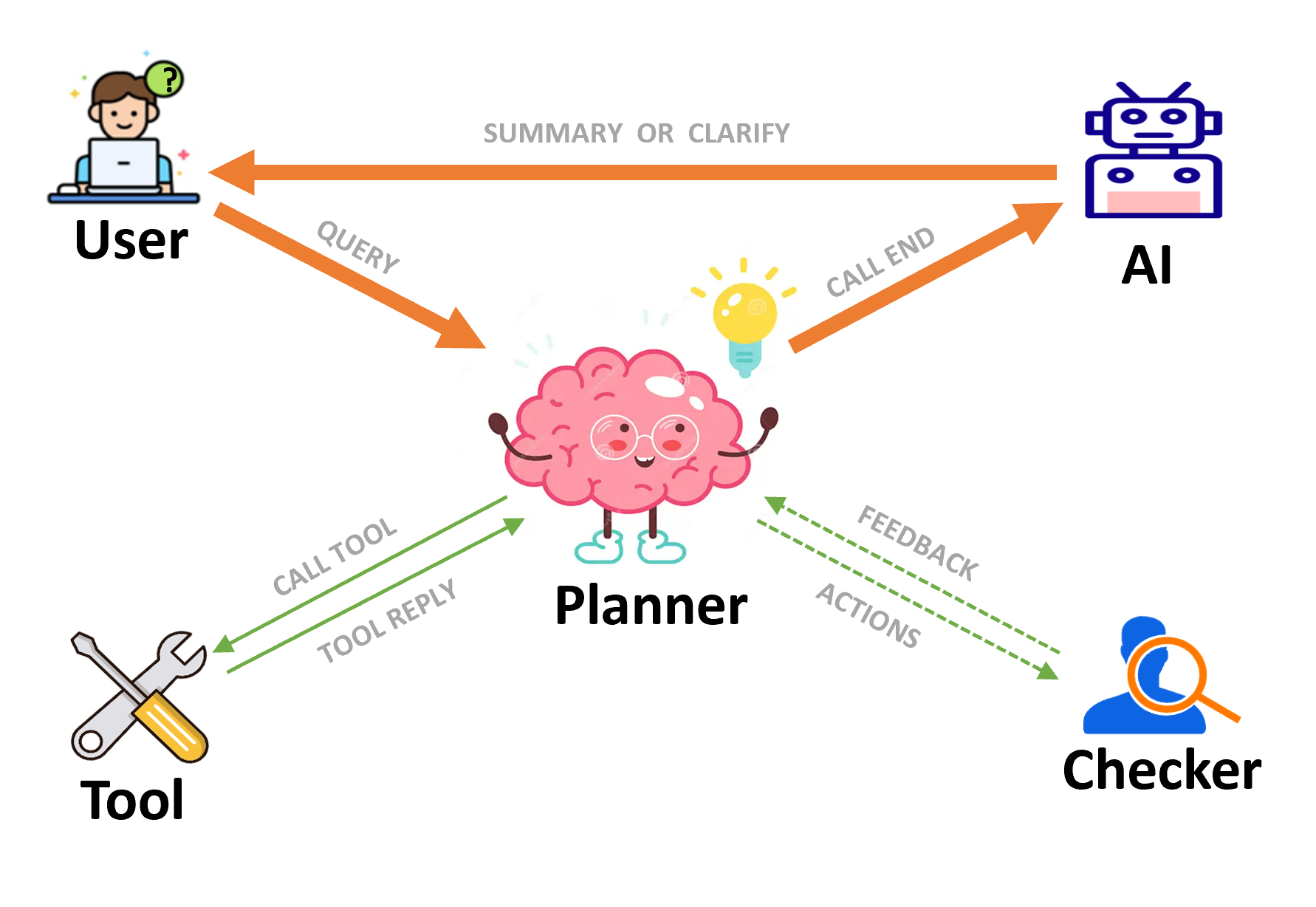} 
  \caption {The multi-agent framework.}
  \label{fig:five_role}
\end{figure}

\section{Benchmark Construction}\label{chap:4}
To construct multi-mission test data, and thoroughly explore the mission switching space, we proposed a novel data generation framework. 
In this section, we explain the proposed framework and how to construct the benchmark. Subsection \ref{chap:4.1} presents the five roles in the framework and their interaction mechanism. Subsection \ref{chap:4.2} describes how these roles complete a mission. It includes specifying mission-types and setting up dependencies with earlier missions for later ones. Subsection  \ref{chap:4.3} we expand the scope from generating a single mission to creating a test data with multiple related missions. Subsequently, we thoroughly explore the mission switching space to construct the entire benchmark. Furthermore, Appendixes \ref{appendix-a} and  \ref{appendix-b} present the method for collecting tools and the distribution of the test set.

\subsection{Data Generation Framework}\label{chap:4.1}

We employ five agents to generate multi-mission test data. We simulate this process with a single LLM. For each dialogue, we assign different roles and specific tasks to the LLM, denoted  $\mathbf{R}$. We define five roles: User, Planner, AI, Checker, and Tool, represented as $R_{user}$, $R_{planner}$, $R_{AI}$, $R_{checker}$, and $R_{tool}$ respectively. The Planner is the key to analyzing the mission, planning tool invocation paths, and deciding action-types. Figure \ref{fig:five_role} shows the interaction among these five roles.

In this framework, only $R_{AI}$ communicates with $R_{user}$, and $R_{planner}$ gets instructions from $R_{user}$. When $R_{planner}$ starts $A_{single}$ or $A_{multi}$, $R_{tool}$ simulates tool feedback. For $A_{clarify}$ or $A_{chat}$, $R_{AI}$ asks about tool parameters or summarizes responses. $R_{checker}$ checks the format and sequence of $R_{planner}$'s plans, ensuring accurate planning. Note that $R_{checker}$ is only involved in data generation. Moreover, $R_{user}$ has different tasks at different stages. $R_{user}^{Q}$ responses to generate a new mission, while $R_{user}^{A}$  responses to answer the questions of $R_{AI}$.

We provide the prompts for the roles mentioned above in Appendix \ref{appendix-c}.

\subsection{Generate Single Mission}\label{chap:4.2}

We first introduce how to construct a single mission using the proposed multi-agent framework. 

In the generation process, we first generate user missions. 
When generating user missions, we first sample a tool list for the missions. 

To achieve a desirable mission type, we insert the predefine action-type $A_i$ into the role prompt $R_{user}^{Q}$.

To generate related missions, we generate several candidate missions, then employ expert refinement to get the final successive mission. We categorize mission relationships into three types: implicit understanding, ellipsis, and long-term memory, and insert the relationship types into $R_{user}^{Q}$ to generate three candidate missions. The $R_{user}^{Q}$ also contains the previous user-AI dialogues. Finally, we manually select and refine the candidate missions to achieve the final one.
 
With the user missions, we use the five roles mentioned above to complete the entire execution.

\begin{figure}
  \centering
  \includegraphics[width=0.35\linewidth]{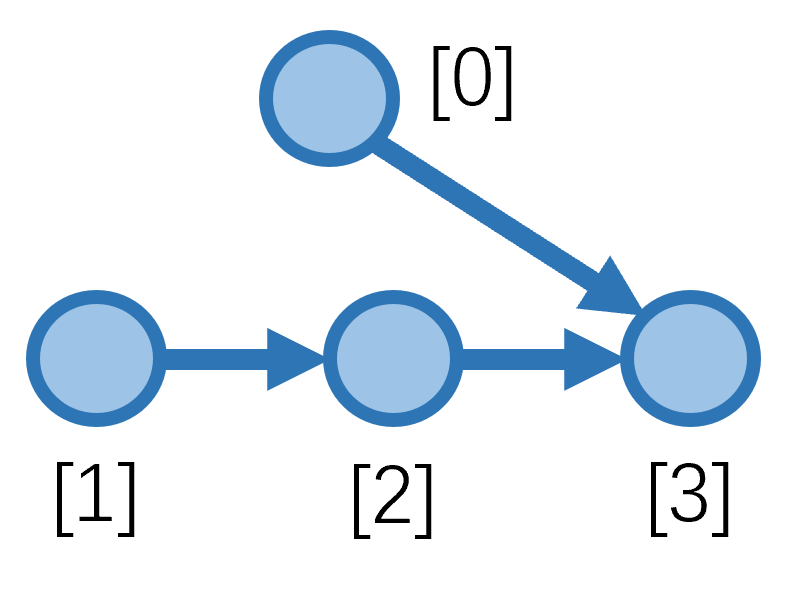} 
  \caption {The dependencies among tools.}
  \label{fig:toy_1}
\end{figure}

\begin{figure*}[h]
  \centering
  \includegraphics[width=0.90\linewidth]{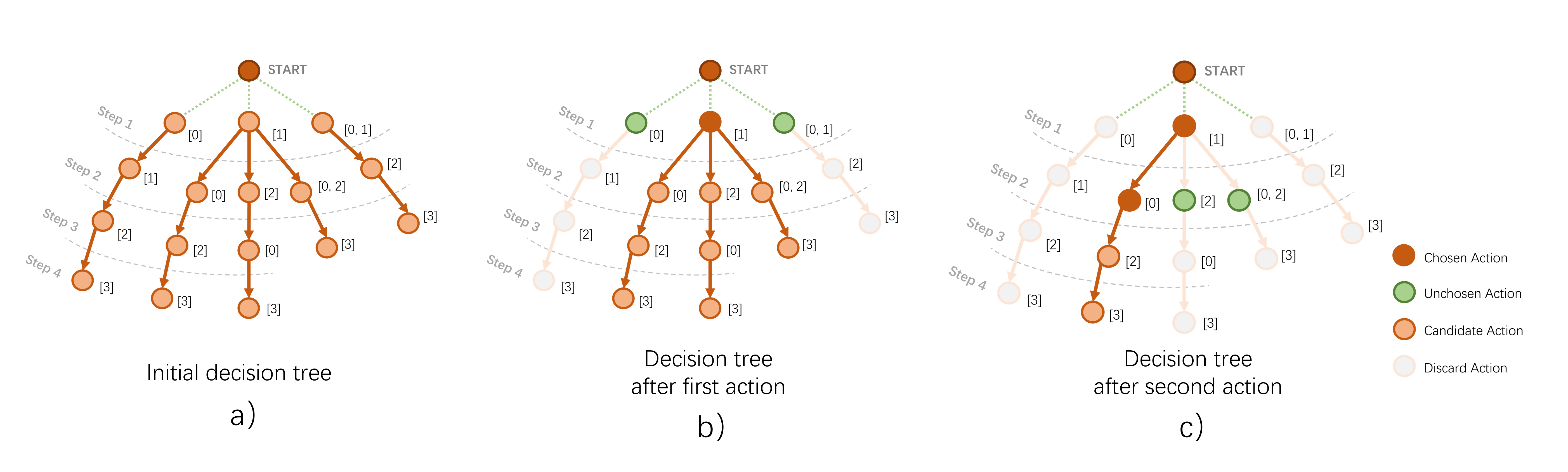} 
  \caption {Visualization of the dynamic decision tree during evaluation.}
  \label{fig:toy_2}
\end{figure*}

\subsection{Construct the Whole Benchmark}\label{chap:4.3}
In Subsection \ref{chap:4.2}, we obtain the ability to generate a specific type of mission and create related missions. Subsequently, we apply this ability to construct the benchmark. This benchmark aims to fully demonstrate the diversity of mission switching in the test data. To achieve this goal, we employ the proposed method to explore the entire mission switching space in prefixed mission number.

First, we identify all combinations of action-types for the given number of missions, represented as $\mathbb{A}={\mathbf{A}_1^1, \mathbf{A}_1^2, ..., \mathbf{A}_N^{4^N}}$. Here, $\mathbf{A}_i^j$ indicates the $j$-th combination for $i$ missions. For $i$ missions, there exist $4^i$ combinations. 

Subsequently, we generate test data independently for each action-type combination. If the action-type combination contains N elements, we use the aforementioned generation framework N times to construct the test data. It is important to note that the generation results from both $R_{tool}$ and $R_{AI}$ are crucial information provided to the agents during our testing process.

\section{Dynamic Evaluation Method}\label{chap:5}

The dependencies among tools lead to multiple possible execution sequences. 
This challenge becomes more pronounced in multi-mission scenarios. To address this, we propose a novel evaluation framework. This framework accurately verifies the correctness and optimality of agent actions. The method follows three steps: dependency analysis, decision tree construction, and path validation.

First, we manually identify tool dependencies. We then implement a topological sorting algorithm with depth-first search to generate all possible execution paths. Unlike previous methods \cite{qiao2024benchmarking, shen2024taskbench}  that produce limited suboptimal paths, our algorithm constructs complete optimal and suboptimal sequences. 

During agent testing, we perform incremental path matching against the decision tree. Each agent action triggers either:
1) Path termination for mismatched actions.
2) Subtree pruning for valid actions, narrowing subsequent options.

To illustrate the process clearly, take a simplified toy example. Consider a user aiming to create a PowerPoint presentation about the year's most popular movie. This task requires four tools: Tool 0 for creating the presentation, Tool 1 for retrieving the popular movie ranking, Tool 2 for gathering detailed movie information, and Tool 3 for transforming this information into slides, labeled as [0], [1], [2], and [3] respectively.

Analysis shows [2] needs parameters from [1], and [3] depends on parameters from [0] and [2]. Figure \ref{fig:toy_1} shows this dependency.Figure \ref{fig:toy_2} a) shows the initial decision tree based on tool dependencies. Here, [0, 1] means tools [0] and [1] are called in parallel. This tree reveals five candidate paths to complete the task with three to four tool calls.

When the agent calls Tool [1] in the first step, check if this action is among the first-step candidate actions. Then, prune the sub-decision trees related to operations [0] and [0,1], getting an updated decision tree as in Figure \ref{fig:toy_2} b). In the second step, when the agent calls Tool [0], confirm the action's correctness and prune the sub-decision trees for candidate actions [0] and [0,2] in the second layer, as in Figure \ref{fig:toy_2} c). At this point, only one candidate path remains, and verify its correctness by sequential path matching.

Additionally, we calculate two metrics. Success rate: percentage of valid paths completed. Optimality rate: percentage of paths that match minimal tool invocations. Appendix \ref{appendix-e} provides formal algorithm specifications.

\section{Experiments}\label{chap:6}
The Multi-Mission Tool Bench consists of 1,024 test entries, each containing one to four missions. We divide the test set into four subsets based on the number of missions, with each subset containing 256 entries.

We evaluated 24 state-of-the-art models on the test set, including closed-source general models, open-source general models, and specialized models. Specifically, the closed-source general models are: o1-2024-12-17\cite{openai-o1}, GPT-4o-2024-11-20\cite{achiam2023gpt}, Gemini-1.5-Pro-002\cite{team2024gemini}, Mistral-Large-2411\cite{mistral-large}, and doubao-1.5-pro-32k\cite{Doubao}. The open-source general models include: Qwen2.5-Instruction-Series\cite{yang2024qwen2}, GLM-4-9B-Chat\cite{glm2024chatglm}, DeepSeek-R1\cite{guo2025deepseek}, DeepSeek-V3\cite{liu2024deepseek}, and Llama-3.3-70B-Instruct\cite{dubey2024llama}. The specialized models are: Toolace \cite{liu2024toolace}, Hammer2.1-Series\cite{lin2024hammer}, watt-tool-8b\cite{shi2024direct}, xLAM-7b-fc-r\cite{zhang2024xlam}, and gorilla-openfunctions-v2\cite{gorilla-openfunctions-v2}. Model sizes range from several hundred billions to 70b, 30b, and the smallest at 0.5b.

This section details the test results and analysis. Subsection \ref{chap:6.1} shows the overall performances. Subsection \ref{chap:6.2} analyzes effects of the number of missions, mission action-types, and mission switching. Subsection \ref{chap:6.3} explores the impact of inter-mission relationship types. Further error analysis is detailed in Appendix \ref{appendix-f}.

\subsection{Overview}\label{chap:6.1}

\begin{figure*}[h]
  \centering
  \includegraphics[width=1.0\linewidth]{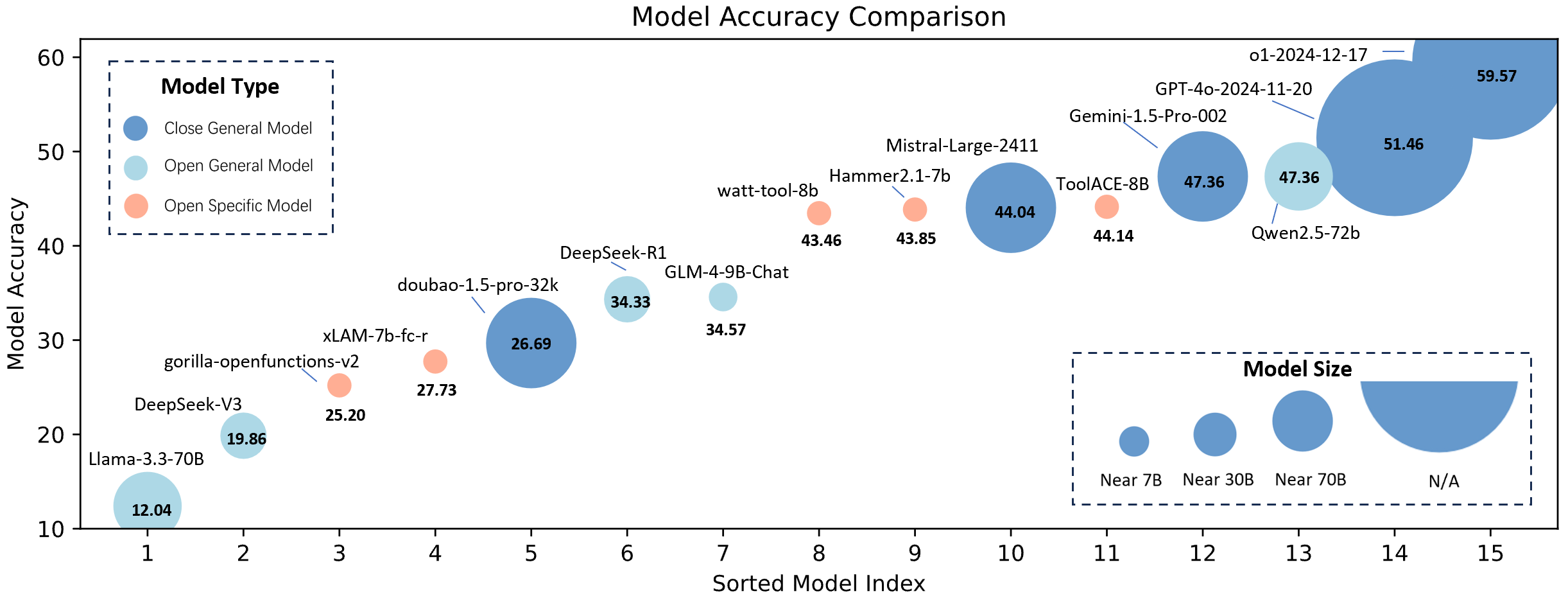} 
  \caption {Overall accuracy of agents on the whole benchmark. }
  \label{fig:exp_1}
\end{figure*}

This subsection analyzes the accuracy of models on the whole dataset, with Figure \ref{fig:exp_1} showing the accuracy of 15 models. The models are arranged from low to high accuracy, with different colored dots indicating model types and varying dot sizes representing model sizes.

From the analysis of Figure \ref{fig:exp_1}, we draw the following conclusions. The o1 model, with strong reasoning capabilities, shows a significant accuracy advantage. Open-source models, such as Qwen-72b, are narrowing the gap with the top close-source models. General models like DeepSeek-V3 and doubao-1.5-pro perform well in other missions but have a clear weakness in tool utilization. Notably, small specialized models like ToolACE achieve comparable performance to large-scale general models.

\begin{figure}[h]
  \centering
  \includegraphics[width=0.90\linewidth]{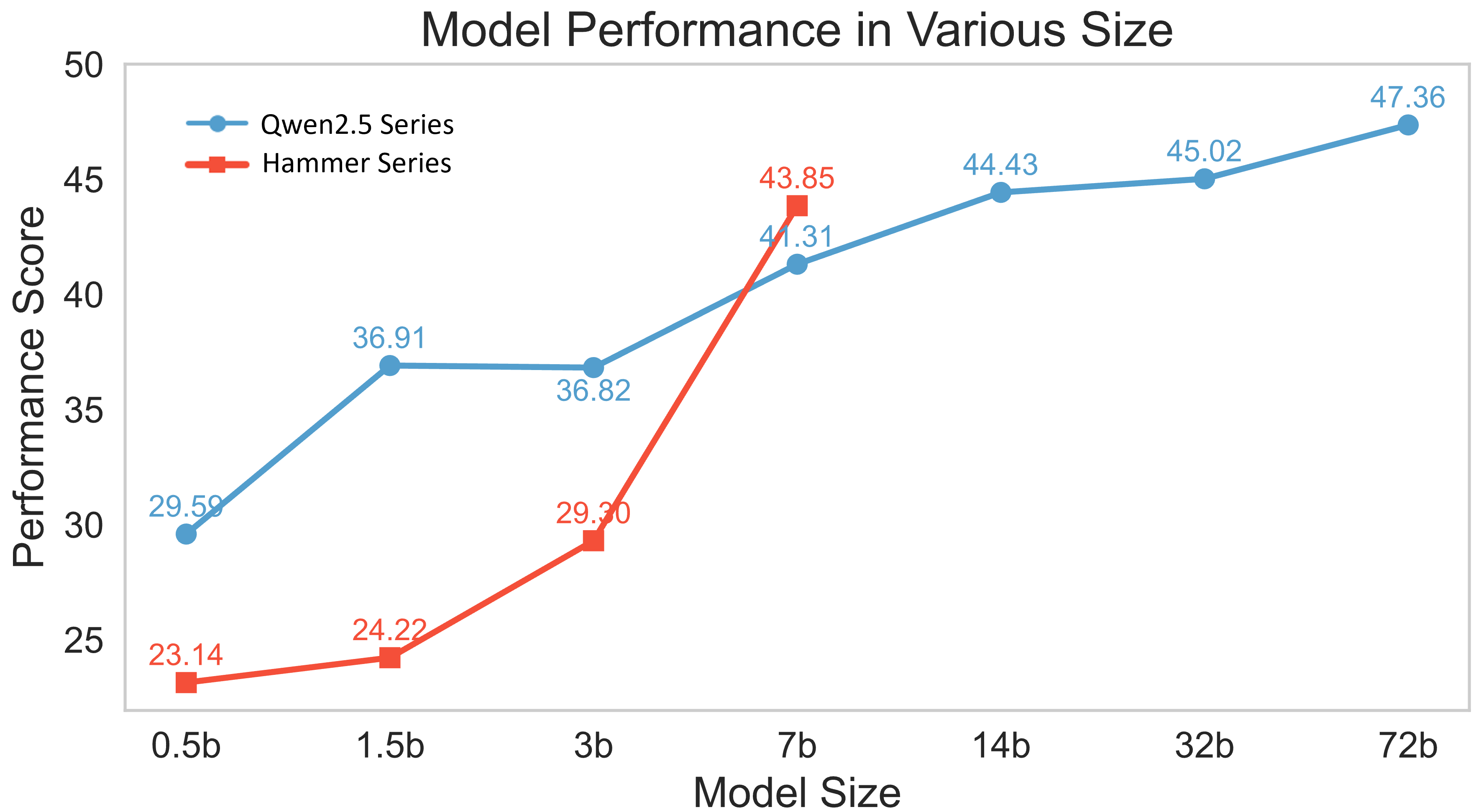} 
  \caption {The performance of two series agents.}
  \label{fig:exp_2}
\end{figure}

Figure \ref{fig:exp_2} illustrates the performance of different scale models in the Qwen2.5-Instruction-Series and Hammer2.1-Series. As shown, there is a positive correlation between model scale and accuracy. Interestingly, specialized models experience a faster decline in accuracy. To explain this phenomenon, more research is needed.

\subsection{Impact of Mission Switching}\label{chap:6.2}

This study examines the impact of mission quantity, mission-type, and mission transition on agent robustness. 

\begin{figure}[h]
  \centering
  \includegraphics[width=0.90\linewidth]{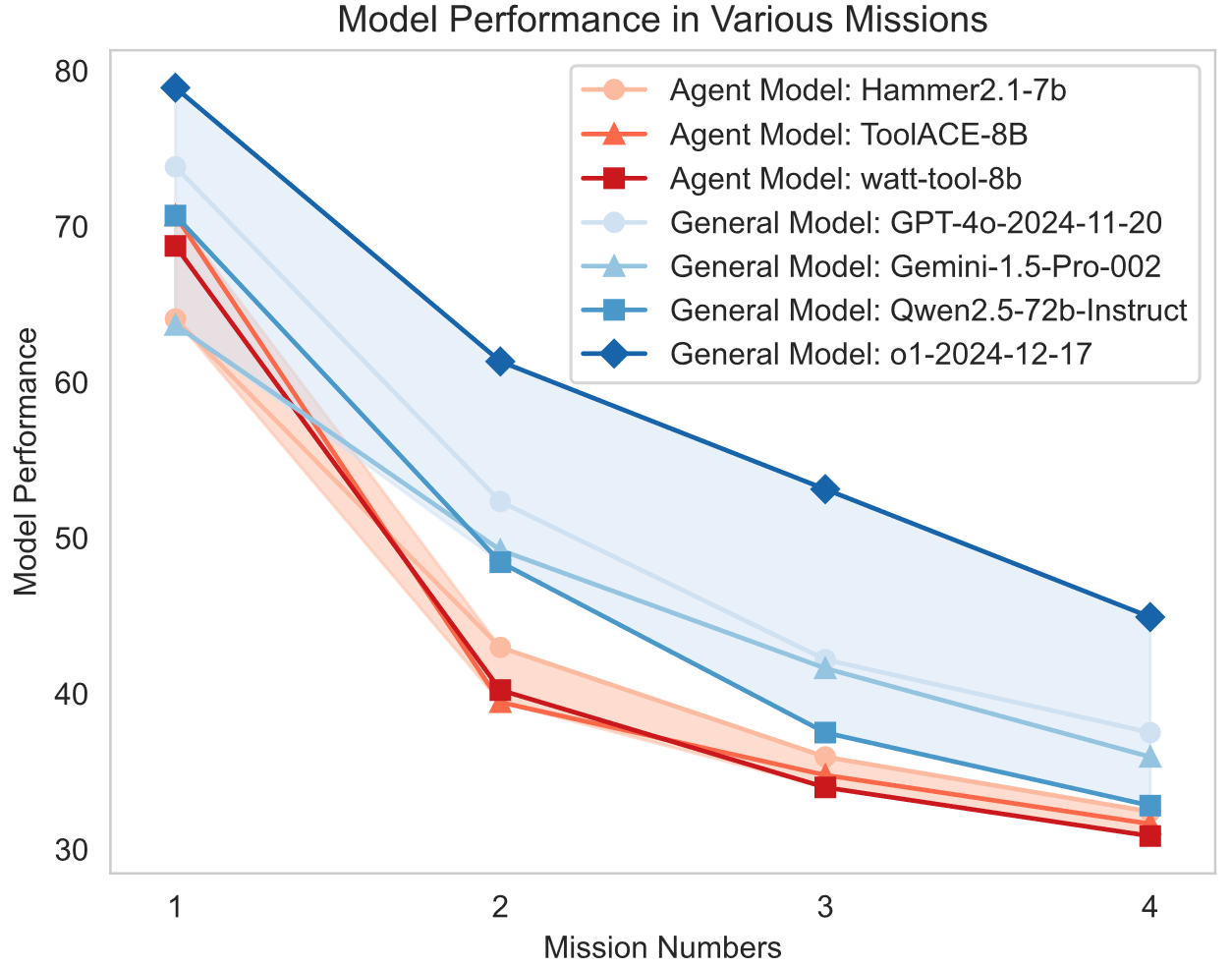} 
  \caption {The impact of various mission number on the agents.}
  \label{fig:exp_3}
\end{figure}

Seven models with better overall performance were selected for detailed analysis, including four general models and three specialized models. Figure \ref{fig:exp_3} presents the performance of these models in various subsets of mission quantities. The results indicate that specialized models perform comparably to stronger general models on single missions but experience a rapid decline in accuracy in multi-mission scenarios. This confirms our hypothesis that current research overlooks the influence of multi-mission. Furthermore, even the most advanced o1 model demonstrates a noticeable decrease in capability when handling multiple missions.

\vspace{1pt}

\begin{figure*}[h]
  \centering
  \includegraphics[width=0.83\linewidth]{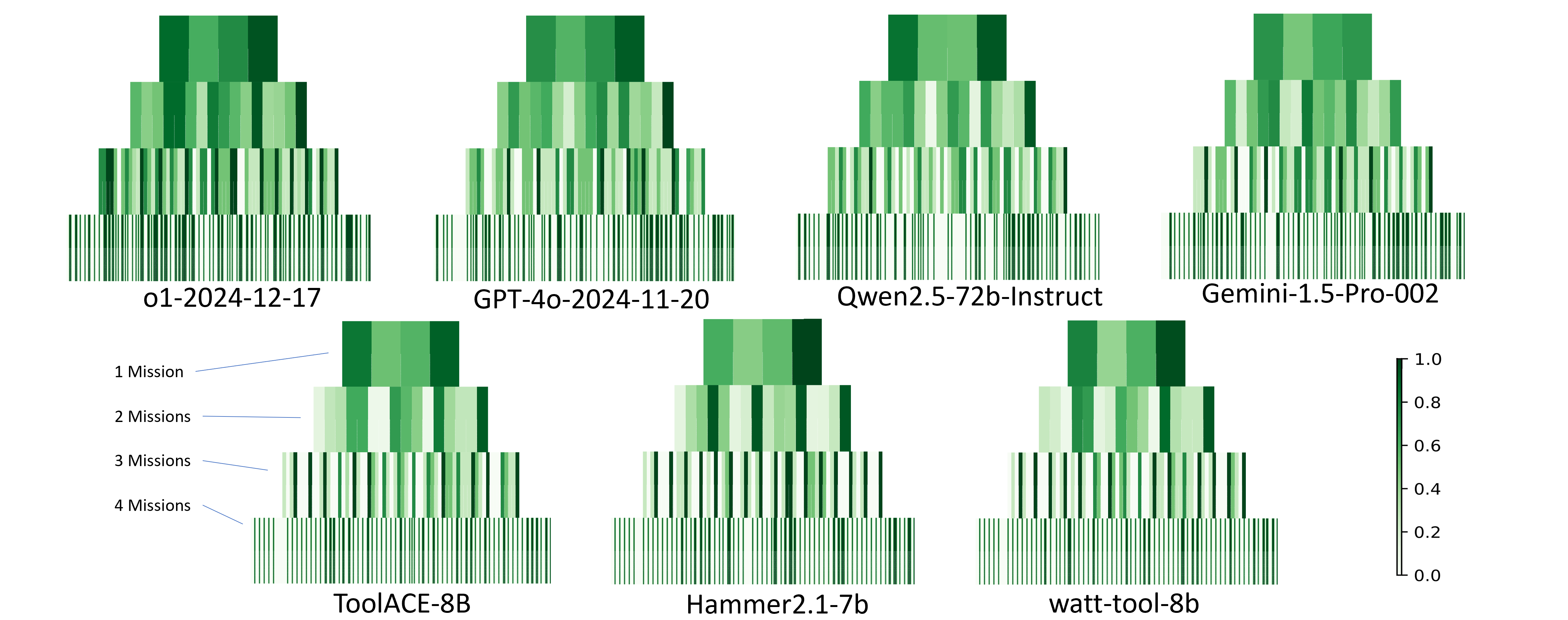} 
  \caption {Visualization of the robustness of agents in the mission switching space. }
  \label{fig:exp_4}
\end{figure*}
\vspace{5pt}

\begin{table*}[h]
    \centering
    \scriptsize  
    \resizebox{0.90\textwidth}{!}{\begin{tabular}{l|cccccc|cc}
\hline 
    \toprule
    \thead{1} Agent& 
    \thead{${A_{\mathrm{single}}}$} & 
    \thead{${A_{\mathrm{chat}}}$} & 
    \thead{${A_{\mathrm{clarity}}}$} & 
    \thead{${A_{\mathrm{multi}}^P}$} & 
    \thead{${A_{\mathrm{multi}}^S}$} & 
    \thead{${A_{\mathrm{multi}}^{S+P}}$} &
    \thead{\scriptsize Optimal \\ \scriptsize Path Rate} & 
    \thead{\scriptsize Accomplished \\ \scriptsize Progress} \\
    \midrule
    \hline
    \tiny o1-2024-12-17$^{\dag}$ & \tiny \cellcolorheat{63.28} & \tiny \cellcolorheat{91.41} & \tiny \cellcolorheat{45.70} & \tiny \cellcolorheat{50.32} & \tiny \cellcolorheat{12.50} & \tiny \cellcolorheat{19.05} & \tiny \cellcolorheatblue{39.42} & \tiny \cellcolorheatblue{30.15} \\
    \tiny GPT-4o-2024-11-20 $^{\dag}$& \tiny \cellcolorheat{54.69} & \tiny \cellcolorheat{74.61} & \tiny \cellcolorheat{35.94} & \tiny \cellcolorheat{51.59} & \tiny \cellcolorheat{18.75} & \tiny \cellcolorheat{23.81} & \tiny \cellcolorheatblue{41.08} & \tiny \cellcolorheatblue{45.56} \\
    \tiny Gemini-1.5-Pro-002$^{\dag}$ & \tiny \cellcolorheat{49.61} & \tiny \cellcolorheat{77.73} & \tiny \cellcolorheat{35.94} & \tiny \cellcolorheat{37.58} & \tiny \cellcolorheat{6.25} & \tiny \cellcolorheat{8.33} & \tiny \cellcolorheatblue{26.14}  & \tiny \cellcolorheatblue{16.58} \\
    \tiny Qwen2.5-72b-Instruct$^{\ddag}$ & \tiny \cellcolorheat{56.25} & \tiny \cellcolorheat{74.61} & \tiny \cellcolorheat{27.34} & \tiny \cellcolorheat{45.22} & \tiny \cellcolorheat{18.75} & \tiny \cellcolorheat{7.14} & \tiny \cellcolorheatblue{30.29} & \tiny \cellcolorheatblue{19.43} \\
    \tiny ToolACE-8B$^{\star}$ & \tiny \cellcolorheat{43.75} & \tiny \cellcolorheat{87.11} & \tiny \cellcolorheat{22.66} & \tiny \cellcolorheat{35.67} & \tiny \cellcolorheat{0.00} & \tiny \cellcolorheat{3.57} & \tiny \cellcolorheatblue{24.07} & \tiny \cellcolorheatblue{9.55} \\
    \tiny Mistral-Large-2411$^{\dag}$ & \tiny \cellcolorheat{57.03} & \tiny \cellcolorheat{55.86} & \tiny \cellcolorheat{31.64} & \tiny \cellcolorheat{41.40} & \tiny \cellcolorheat{12.50} & \tiny \cellcolorheat{16.67} & \tiny \cellcolorheatblue{29.88} & \tiny \cellcolorheatblue{37.69} \\
    \tiny Hammer2.1-7b$^{\star}$ & \tiny \cellcolorheat{ 28.13 } &  \tiny \cellcolorheat{ 91.27 } & \tiny \cellcolorheat{ 31.25 } & \tiny \cellcolorheat{ 28.03 } & \tiny \cellcolorheat{ 6.25 } & \tiny \cellcolorheat{ 3.57 } & \tiny \cellcolorheatblue{19.67} & \tiny \cellcolorheatblue{9.72} \\
    \tiny watt-tool-8b$^{\star}$ & \tiny \cellcolorheat{40.63} & \tiny \cellcolorheat{91.80} & \tiny \cellcolorheat{23.05} & \tiny \cellcolorheat{29.94} & \tiny \cellcolorheat{0.00} & \tiny \cellcolorheat{0.00} & \tiny \cellcolorheatblue{19.50} & \tiny \cellcolorheatblue{8.38} \\
    \tiny GLM-4-9B-Chat$^{\ddag}$ & \tiny \cellcolorheat{30.08} & \tiny \cellcolorheat{89.84} & \tiny \cellcolorheat{10.16} & \tiny \cellcolorheat{12.10} & \tiny \cellcolorheat{12.50} & \tiny \cellcolorheat{0.00} & \tiny \cellcolorheatblue{0.00} & \tiny \cellcolorheatblue{12.23} \\
    \tiny DeepSeek-R1$^{\ddag}$ & \tiny \cellcolorheat{27.50} & \tiny \cellcolorheat{68.27} & \tiny \cellcolorheat{13.39} & \tiny \cellcolorheat{44.19} & \tiny \cellcolorheat{33.33} & \tiny \cellcolorheat{6.06} & \tiny \cellcolorheatblue{33.61} & \tiny \cellcolorheatblue{39.17} \\
    \tiny doubao-1.5-pro-32k$^{\dag}$ & \tiny \cellcolorheat{60.16} & \tiny \cellcolorheat{25.78} & \tiny \cellcolorheat{5.86} & \tiny \cellcolorheat{36.94} & \tiny \cellcolorheat{18.75} & \tiny \cellcolorheat{9.52} & \tiny \cellcolorheatblue{5.39} & \tiny \cellcolorheatblue{38.53} \\
    \tiny xLAM-7b-fc-r$^{\star}$ & \tiny \cellcolorheat{14.45} & \tiny \cellcolorheat{86.33} & \tiny \cellcolorheat{5.08} & \tiny \cellcolorheat{7.64} & \tiny \cellcolorheat{0.00} & \tiny \cellcolorheat{1.19} & \tiny \cellcolorheatblue{4.56} & \tiny  \cellcolorheatblue{9.55} \\
    \tiny gorilla-openfunctions-v2$^{\star}$ & \tiny \cellcolorheat{2.34} & \tiny \cellcolorheat{90.63} & \tiny \cellcolorheat{4.30} & \tiny \cellcolorheat{5.73} & \tiny \cellcolorheat{0.00} & \tiny \cellcolorheat{0.00} & \tiny \cellcolorheatblue{3.73} & \tiny \cellcolorheatblue{4.86} \\
    \tiny DeepSeek-V3$^{\ddag}$ & \tiny \cellcolorheat{22.09} & \tiny \cellcolorheat{41.58} & \tiny \cellcolorheat{7.51} & \tiny \cellcolorheat{4.81} & \tiny \cellcolorheat{0.00} & \tiny \cellcolorheat{4.55} & \tiny \cellcolorheatblue{4.05} & \tiny \cellcolorheatblue{24.13} \\
    \tiny Llama-3.3-70B-Instruct$^{\ddag}$ & \tiny \cellcolorheat{29.30} & \tiny \cellcolorheat{19.92} & \tiny \cellcolorheat{0.00} & \tiny \cellcolorheat{0.64} & \tiny \cellcolorheat{0.00} & \tiny \cellcolorheat{0.00} & \tiny \cellcolorheatblue{0.00} & \tiny \cellcolorheatblue{12.40} \\    \hline 
    \bottomrule
    \end{tabular}
    }
    \caption{The performance of agents in various type of missions, and the quantitative evaluation results on $A_{multi}$ missions. Here, $\dag$ and  $\ddag$  represent  close-source and open-source general model, $\star$ represents specific model. }
    \label{tab:exp_1}
\end{table*}

\vspace{1pt}

We further analyze the performance of the seven models across different action-type combinations. Following the structure of Figure \ref{fig:instruction} b), in Figure \ref{fig:exp_4}, we visualize the models' performance in the action-type space with heatmaps. Each heatmap pyramid represents a model's performance, with each layer corresponding to a sub-testset and its action-type combinations. Deeper colors signify higher accuracy. Greater color contrast within the same layer, with a larger proportion of lighter areas, indicates poorer robustness of the model. The findings reveal that the best performing o1 model also exhibits the highest robustness. In contrast, the three specialized models show less stability than the general models.

\vspace{1pt}

\subsection{Impact of Mission Types}\label{chap:6.3}

Moreover, we divide the test set by mission action-type and analyze the performance of all models, as shown in Table \ref{tab:exp_1}. The heatmap reveals several observations: models exhibit varying strengths and weaknesses across different action-types. For instance, most models struggle to determine whether the necessary parameters are missing($A_{clarity}$). Although many models have the ability to handle $A_{multi}$ missions, they still face challenges in handling complex scenarios such as tackling $A_{multi}^S$ and $A_{multi}^{S+P}$ missions.

For multi-tool invocation, we introduce two new metrics, with results displayed on the far right of Table \ref{tab:exp_1}. The first is the optimal path rate, where the general models perform notably well. Additionally, instead of using hard labels to indicate mission success, we propose accomplished progress metric to assess model capability.

\subsection{Impact of Related Mission}\label{chap:6.4}

This subsection examines how mission relationship types affect agent performance. As mentioned, all subsequent missions in our benchmark are closely relate to preceding missions, and we have abstracted three types of mission relationships.

Table \ref{tab:exp_2} presents the accuracy of all models in the three types of mission relationship. Long-term memory has the most significant impact on model performance, followed by the absence of core components in the problem( ellipsis ).

\begin{table}[h]

        \centering
    \scriptsize 
        \begin{tabular}{l|ccc}
        \hline
        \makecell[c]{\textbf{Agent}}  & \makecell[c]{\textbf{Implicit}} & \makecell[c]{\textbf{Ellipsis}} & \makecell[c]{\textbf{Long-Term}}\\
        \hline
               o1-2024-12-17$^{\dag}$ & \cellcolorheat{57.31} & \cellcolorheat{54.17} & \cellcolorheat{43.58} \\
GPT-4o-2024-11-20$^{\dag}$ & \cellcolorheat{42.69} & \cellcolorheat{52.92} & \cellcolorheat{34.64} \\
Gemini-1.5-Pro-002$^{\dag}$ & \cellcolorheat{46.99} & \cellcolorheat{42.08} & \cellcolorheat{31.84} \\
Qwen2.5-72b-Instruct$^{\ddag}$ & \cellcolorheat{40.11} & \cellcolorheat{47.08} & \cellcolorheat{28.49} \\
ToolACE-8B$^{\star}$ & \cellcolorheat{38.68} & \cellcolorheat{35.83} & \cellcolorheat{27.93} \\
Mistral-Large-2411$^{\dag}$ & \cellcolorheat{35.24} & \cellcolorheat{39.17} & \cellcolorheat{30.17} \\
Hammer2.1-7b$^{\star}$ & \cellcolorheat{43.55} & \cellcolorheat{34.58} & \cellcolorheat{27.93} \\
watt-tool-8b$^{\star}$ & \cellcolorheat{40.97} & \cellcolorheat{32.92} & \cellcolorheat{26.26} \\
GLM-4-9B-Chat$^{\ddag}$ & \cellcolorheat{35.82} & \cellcolorheat{26.25} & \cellcolorheat{21.23} \\
DeepSeek-R1$^{\ddag}$ & \cellcolorheat{30.06} & \cellcolorheat{32.28} & \cellcolorheat{18.67} \\
doubao-1.5-pro-32k$^{\dag}$ & \cellcolorheat{25.79} & \cellcolorheat{28.33} & \cellcolorheat{22.91} \\
xLAM-7b-fc-r$^{\star}$ & \cellcolorheat{30.37} & \cellcolorheat{22.92} & \cellcolorheat{19.55} \\
gorilla-openfunctions-v2$^{\star}$ & \cellcolorheat{29.80} & \cellcolorheat{21.67} & \cellcolorheat{16.20} \\
DeepSeek-V3$^{\ddag}$ & \cellcolorheat{17.28} & \cellcolorheat{18.07} & \cellcolorheat{13.39} \\
Llama-3.3-70B-Instruct$^{\ddag}$ & \cellcolorheat{9.17} & \cellcolorheat{13.33} & \cellcolorheat{11.17} \\
                \hline
        \end{tabular}
            \caption{The impact of mission relation types on agent performance. }
\label{tab:exp_2}
        \end{table}

\section{Conclusion}
This paper introduces a novel multi-mission benchmark to evaluate the robustness of LLM-based agents. Evaluations reveal that current agents exhibit varying degrees of limitations when addressing multi-mission scenarios. Notably, while specialized agents achieve comparable overall accuracy and single-mission performance to general agents, a significant robustness gap emerges in multi-mission contexts.
Moreover, all agents struggle with complex multi-tool invocation missions and have shortcomings in related mission handling. We believe that these findings offer valuable insights for guiding future research on the development of LLM-agents.

\section*{Limitations}
In evaluating LLM-based agents from a multi-mission perspective, we identify specific limitations in both mission duration and the data generation framework.

Firstly, our study aims to enhance the diversity of test data in terms of mission variation, yet the diversity in the number of missions remains limited. Specifically, our test data comprises up to four missions. This limitation arises because the mission switching space expands exponentially with an increase in the number of missions, leading to a rapid enlargement of the test set size and additional workload. Moreover, we observe a swift decline in the precision of the model's output as the number of missions increases, indicating that there is no immediate need to explore the model's performance across a larger number of missions.

Secondly, the proposed data generation framework employs multiple iterations and human intervention to ensure the quality of multi-turn dialogue production. This approach suffers the limitations of LLMs in accurately following instructions.

In summary, these limitations emphasize the need for ongoing development in the field of LLM based evaluations.


\appendix

\section{Diverse Toolset Construction}\label{appendix-a}

We generate the toolset based on tool descriptions from public-apis, following the ToolAlpaca approch. This API repository contains 400 tool lists, corresponding to 1600 tools in 50 categories.

In contrast to ToolAlpaca, our approach includes three strategies to enhance tool accuracy and parameter variety. Initially, we utilize LLMs like GPT to refine tool descriptions, addressing the common issue of the absence of constraint parameters in generated tools. For instance, a tool description for querying Spanish weather does not mention Spain in any of its three specific functions, leading to the generated tool cannot validate the query location. Second, we expand parameter types to include complex data structures such as enumerations, arrays, and objects, aligning better with real-world scenarios. Finally, five LLM agent experts review the generated tools. These steps ensure the tools' accuracy and parameter diversity.

\section{Analysis of the Test Data}\label{appendix-b}

\begin{figure}[t]
  \centering
  \includegraphics[width=0.90\linewidth]{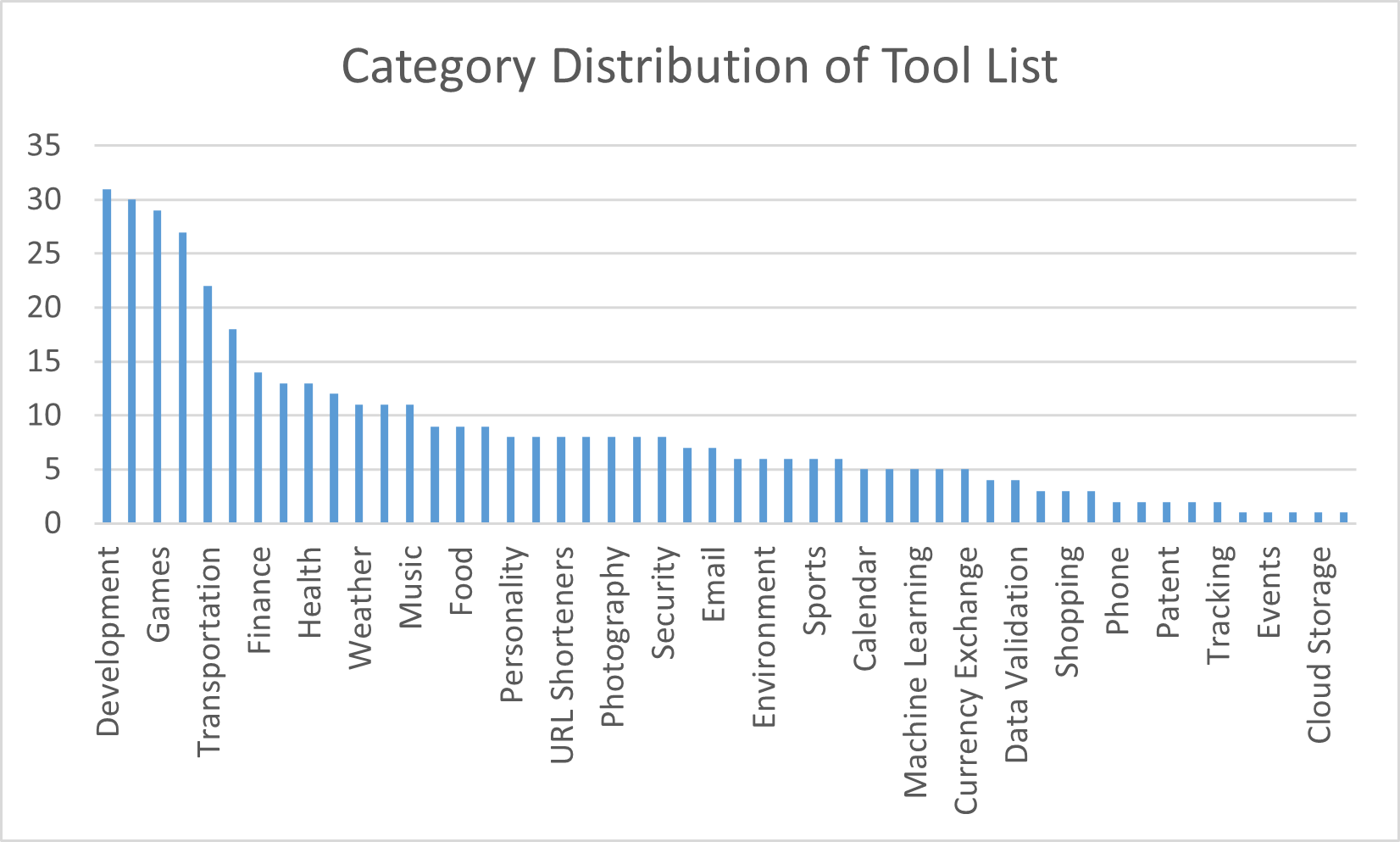} 
  \caption {Category distribution of tools.}
  \label{fig:apdx_1}
\end{figure}

\begin{figure}[h]
  \centering
  \includegraphics[width=0.90\linewidth]{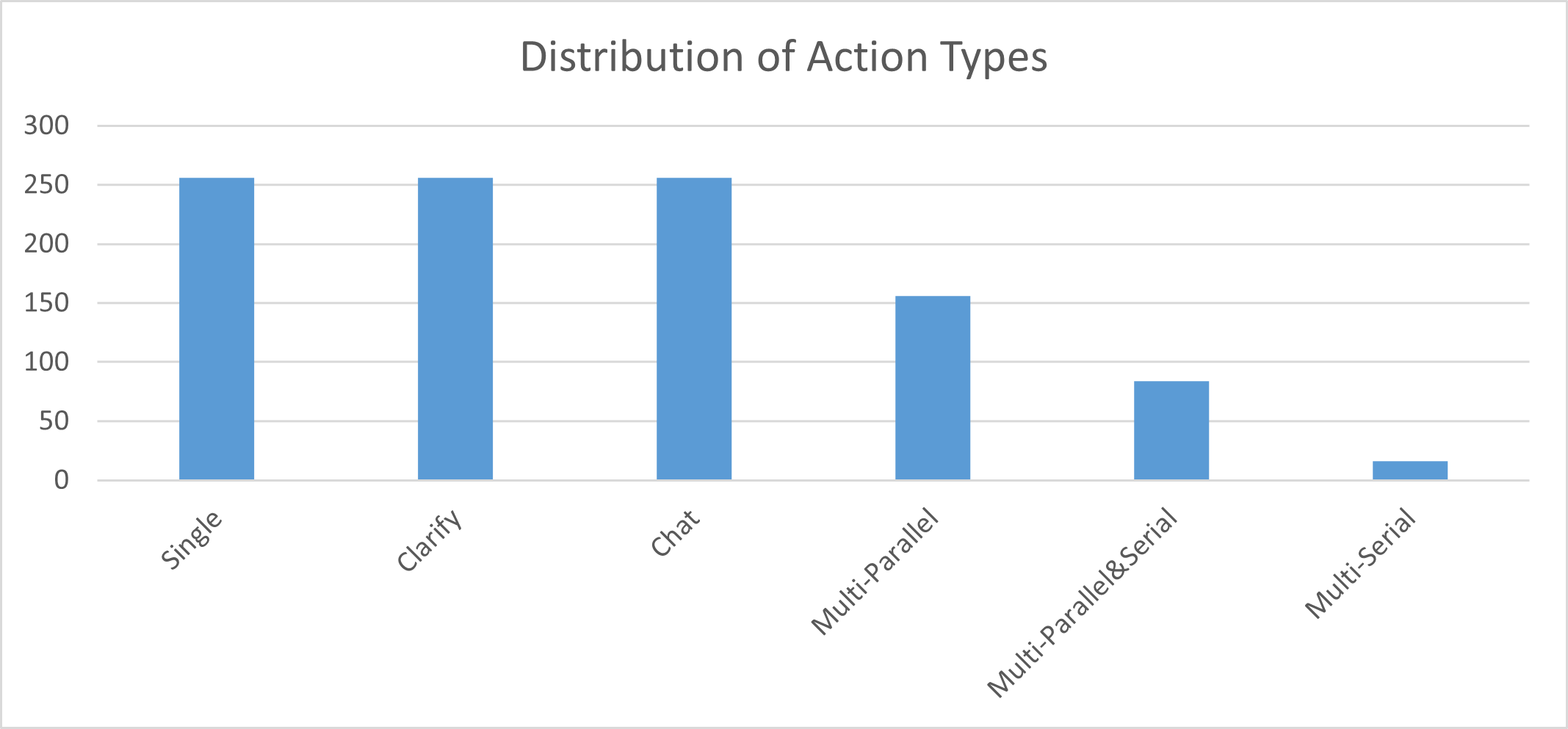} 
  \caption {Distribution of action-types.}
  \label{fig:apdx_2}
\end{figure}

\begin{figure}[h]
  \centering
  \includegraphics[width=0.7\linewidth]{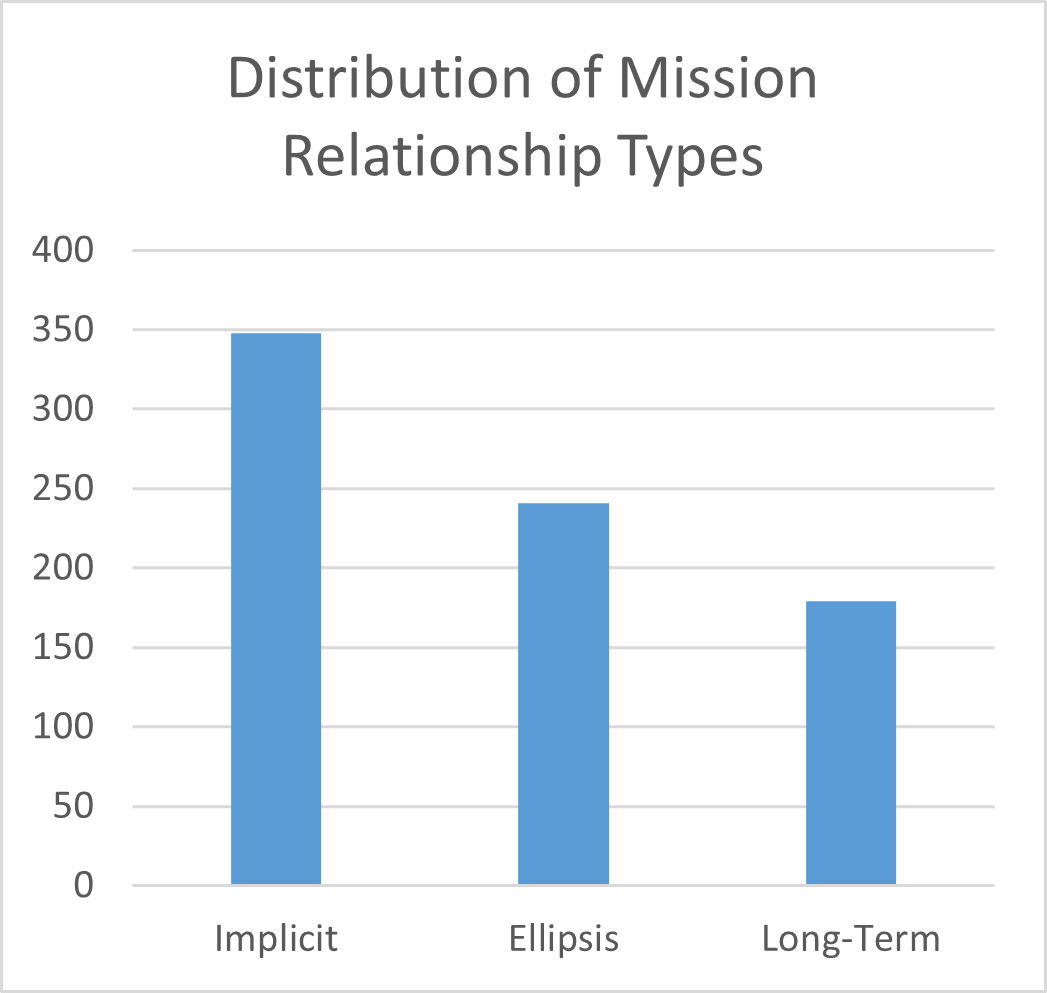} 
  \caption {Distribution of three mission relationship types.}
  \label{fig:apdx_3}
\end{figure}

Figure \ref{fig:apdx_1}, \ref{fig:apdx_2} and \ref{fig:apdx_3} present the proposed dataset from the following three perspectives.

\subsection{Data Examples}
\begin{figure*}[h]
  \centering
  \includegraphics[width=0.90\linewidth]{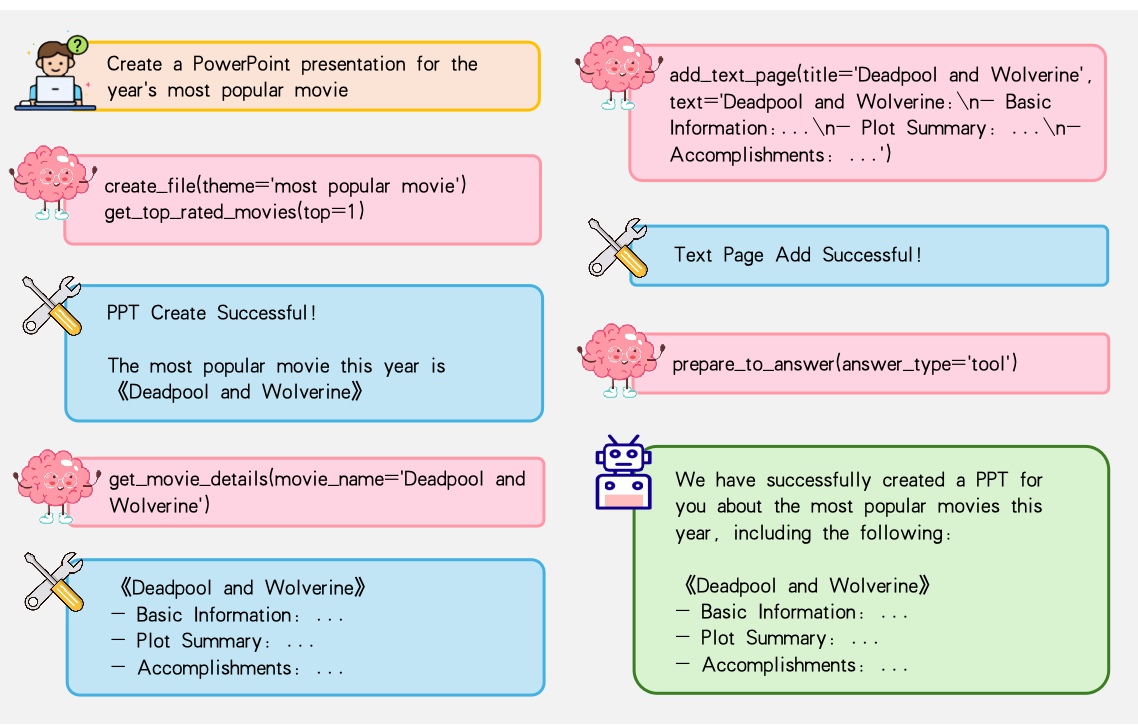} 
  \caption {An Optimal Path Example.}
  \label{fig:apdx_4}
\end{figure*}

\begin{figure*}[h]
  \centering
  \includegraphics[width=0.90\linewidth]{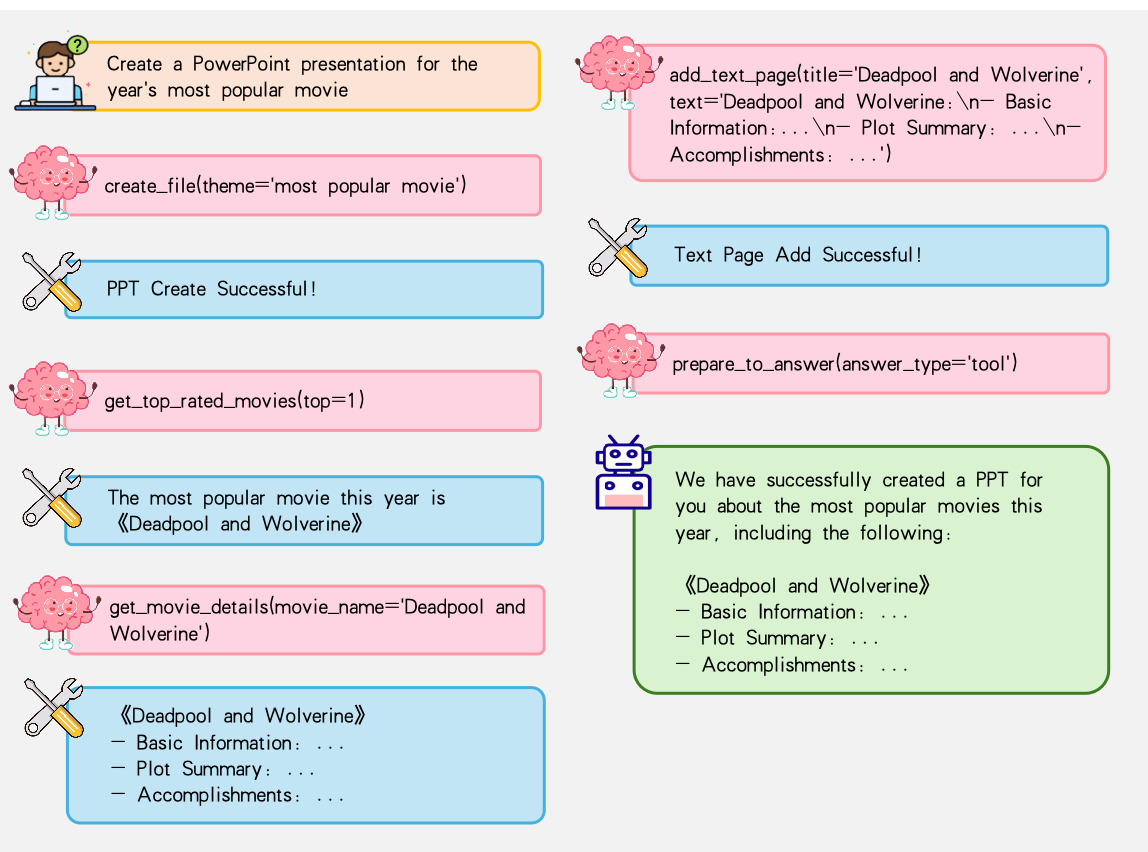} 
  \caption {A Suboptimal Path Example.}
  \label{fig:apdx_5}
\end{figure*}

We present two more examples of mission execution corresponding to the examples in Section  5. Figure \ref{fig:apdx_4} illustrates the execution of the optimal path, while Figure  \ref{fig:apdx_5} shows a non-optimal path execution.

\section{Details of Proposed Evaluation Method}\label{appendix-e}

1. Initialize graph G, indegree table, visitation table, current path, and all paths.

2. Perform topological sorting and depth-first traversal based on parallel combination and permutation.

    2.1 For each search, find all nodes with an indegree of 0 and arrange all possible combinations based on the number of nodes. Specifically, since nodes with an indegree of 0 are independent, they can be combined arbitrarily. When the number of nodes in a combination is greater than 1, it indicates that these nodes can be called in parallel. It is this method that allows our algorithm to enumerate all possible paths, including parallel and serial-parallel calls, as opposed to being limited to serial calls only, compared to naive topological sorting.

    2.2 Traverse each combination, add the combination to the current path, and update the indegree and visitation tables.
    
    2.3 Continue with depth-first traversal until the number of nodes in the path matches the number of nodes in the annotated answer, completing the generation of one path, and add it to all paths.
    
    2.4 Repeat the above steps until the traversal is complete.

3. Based on the path length, divide into the optimal path and the suboptimal path.

\section{Further Analysis of Agent Performance}\label{appendix-f}

In addition to the analytical perspectives mentioned in the main text, we analyze the error types of the agents.

We categorize errors into tool errors and parameter errors. Specifically, we further divide parameter errors into parameter name hallucinations, parameter value hallucinations, and parameter value errors. Table \ref{tab:apdx_1} lists these error classifications. Stronger agents show a relatively lower proportion of tool errors. Although parameter name hallucinations occur less frequently, they are serious and widespread. The most common parameter error occurs when the agent extracts parameter values.

\begin{table}[h]
    \centering
    \caption{The distribution of agent errors. Here, `Hallu.' is short for hallucination. }
    \scriptsize  
    \begin{tabular}{l|c|c|c|c}
        \hline
        \multirow{3}{*}{\textbf{Agent}} & \multirow{3}{*}{\textbf{\makecell{Tool \\ Errors}}} & \multicolumn{3}{c}{\textbf{Parameter Errors}} \\
        \cline{3-5}
        & & \makecell{Name \\ Hallu.} & \makecell{Value \\  Hallu. } & \makecell{Value \\ Err} \\
        \hline
        o1-2024-12-17 & 83.33 & \cellcolorheatblueblue{0.24} & \cellcolorheatblueblue{5.07} & \cellcolorheatblueblue{11.36} \\
GPT-4o-2024-11-20 & 75.87 & \cellcolorheatblueblue{0.20} & \cellcolorheatblueblue{8.05} & \cellcolorheatblueblue{15.49} \\
Gemini-1.5-Pro-002 & 85.15 & \cellcolorheatblueblue{0.19} & \cellcolorheatblueblue{3.34} & \cellcolorheatblueblue{11.32} \\
Qwen2.5-72b-Instruct & 80.90 & \cellcolorheatblueblue{0.37} & \cellcolorheatblueblue{6.31} & \cellcolorheatblueblue{12.43} \\
ToolACE-8B & 90.56 & \cellcolorheatblueblue{0.17} & \cellcolorheatblueblue{1.75} & \cellcolorheatblueblue{7.52} \\
Mistral-Large-2411 & 78.19 & \cellcolorheatblueblue{0.35} & \cellcolorheatblueblue{6.46} & \cellcolorheatblueblue{15.01} \\
watt-tool-8b & 90.68 & \cellcolorheatblueblue{0.17} & \cellcolorheatblueblue{3.63} & \cellcolorheatblueblue{5.53} \\
GLM-4-9B-Chat & 92.99 & \cellcolorheatblueblue{0.15} & \cellcolorheatblueblue{2.99} & \cellcolorheatblueblue{3.88} \\
DeepSeek-R1 & 95.77 & \cellcolorheatblueblue{0.00} & \cellcolorheatblueblue{2.11} & \cellcolorheatblueblue{2.11} \\
doubao-1.5-pro-32k & 82.35 & \cellcolorheatblueblue{0.28} & \cellcolorheatblueblue{10.69} & \cellcolorheatblueblue{6.67} \\
xLAM-7b-fc-r & 96.36 & \cellcolorheatblueblue{0.27} & \cellcolorheatblueblue{1.35} & \cellcolorheatblueblue{1.89} \\
gorilla-openfunctions-v2 & 98.83 & \cellcolorheatblueblue{0.00} & \cellcolorheatblueblue{0.26} & \cellcolorheatblueblue{0.90} \\
DeepSeek-V3 & 96.57 & \cellcolorheatblueblue{0.00} & \cellcolorheatblueblue{0.90} & \cellcolorheatblueblue{2.53} \\
Llama-3.3-70B-Instruct & 90.53 & \cellcolorheatblueblue{0.33} & \cellcolorheatblueblue{2.45} & \cellcolorheatblueblue{6.69} \\
\hline
    \end{tabular}
\label{tab:apdx_1}
\end{table}

\section{Part Roles Prompt of Agents}\label{appendix-c}

\subsection{Role Prompt of Mission Generation}
We show the role prompt of single mission generation in Figure \ref{P-mission}.

\subsection{Role Prompt of Planner}
We show the role prompt of Planner in Figures \ref{P-planner_1}, \ref{P-planner_2}, \ref{P-planner_3}, \ref{P-planner_4}, \ref{P-planner_5}, \ref{P-planner_6} and  \ref{P-planner_7}.

\subsection{Role Prompt of Tool}

We show the role prompt of Tool in Figures \ref{P-tool}.

\subsection{Role Prompt of AI}

We show the role prompt of AI in Figures \ref{P-AI}.

\begin{figure*}[htbp]
\begin{tcolorbox}[title={Single Tool Invocation Mission Generation Prompt.}]

Please act as a user interacting with a super intelligent agent.

\medskip

This super intelligent agent has access to a range of external tools and can use these tools to solve the missions you propose.

\medskip

Next, please propose 5 missions that you need the super intelligent agent to solve based on the

\medskip

All 5 missions must require the use of $\{\{\{tool\}\}\}$ from the [Tool List] to be completed, and each mission should only require a single call to $\{\{\{tool\}\}\}$.

\medskip

The missions should be specific and diverse.

\medskip

Finally, please output the final result according to the [Format] without generating any extra text.

\medskip

The required parameters for tool $\{\{\{tool\}\}\}$ are: $\{\{\{tool\_required\}\}\}$, and the optional parameters are: $\{\{\{tool\_no\_required\}\}\}$.

\medskip

\medskip

[Requirements]="""

1. The description of the user's mission must include information on all the required parameters needed to call $\{\{\{tool\}\}\}$. For other optional parameters, please add them as you see fit, using natural language.

2. The user's missions should use different types of sentence structures: imperative, declarative, interrogative, etc.

3. The user's missions should include different tones: colloquial, formal, polite, direct, etc.

4. Ensure that the length of the user's missions varies, gradually increasing from short to long.

5. Ensure that the user's missions involve different themes/instances, different scenarios, and different roles.

6. Extract common entities that appear in all descriptions from the [Tool List] and ensure that these entities appear in the user's missions.

7. Do not explicitly specify the tool $\{\{\{tool\}\}\}$ in the user's missions.

"""

\medskip

\medskip

[Tool List]="""

$\{\{\{tool\}\}\}$

"""

\medskip

\medskip

[Format]="""

\{

    \parindent=2em
    
    "mission 1": "xxx",
    
    "mission 2": "xxx",
    
    "mission 3": "xxx",
    
    "mission 4": "xxx",
    
    "mission 5": "xxx",
    
    \noindent
\}

\noindent
"""

\end{tcolorbox}
\caption{Single Tool Invocation Mission Generation Prompt.}
\label{P-mission}
\end{figure*}

\begin{figure*}[htbp]
\begin{tcolorbox}[title={Planner Decision Generation Prompt Part-1.}]

Please act as a Planner within a super intelligent agent.

\medskip

You have access to a series of external tools, and you can solve user missions by invoking these external tools, as detailed in the [Tool List].

\medskip

You are responsible for assessing the completion status of the current user mission and providing thoughts, plans, and actions to be executed.

\medskip

If the Checker\_Planner indicates `no' for correct, it means there is an issue with the decision you made in the previous round. In this case, you should regenerate your decision based on the analysis provided by the Checker\_Planner.

\medskip

However, please be mindful not to include explanations of previously generated incorrect results in your Thoughts!

\medskip

In your Plan, be sure not to mention the use of the prepare\_to\_answer tool and the ask\_user\_for\_required\_parameters tool. Instead, describe these actions in natural language, as the prepare\_to\_answer and ask\_user\_for\_required\_parameters tools are not to be exposed.

\medskip

Refer to the [Planner Output Format] for the output format.

\medskip

[Environmental Information]="""

\{\{\{env\_info\}\}\}

"""

\end{tcolorbox}
\caption{Planner Decision Generation Prompt Part-1.}
\label{P-planner_1}

\end{figure*}

\begin{figure*}[htbp]

\begin{tcolorbox}[title={Planner Decision Generation Prompt Part-2.}]

[Planner Output Format]="""

Planner:

\{

    \parindent=2em
    
    "Mission\_Finish": "Whether the user mission is completed, fill in `yes' if completed, `no' if not completed",
    
    "Thought": "Based on the [Requirements] and [Environmental Information], follow the steps below to give the internal thought process when solving the user mission. You must provide an analysis of the required and optional parameters for each tool that needs to be called.
    
First step, decompose the mission, first analyze whether a tool needs to be called to complete it, and whether there is a suitable tool in the [Tool List].

If a tool needs to be called, which tool(s) should be used to complete the user mission, whether one or multiple tools should be called.

If multiple tools are involved, please provide an analysis of the serial and parallel nature of multiple tools.

Second step, provide an analysis of the required and optional parameters for the first tool that needs to be called (now), in the following order.

1. First, list the required and optional parameters for each tool that needs to be called.

2. Based on the context and user mission, analyze the required parameters, check which information for each tool's required parameters is provided, and explain which are provided and which are missing to ask the user.

3. Finally, analyze the optional parameters. If the user has provided information for optional parameters, briefly explain the situation; otherwise, there is no need to explain.

Note:

1. The analysis process should not be too lengthy; it needs to be concise and clear.

2. Do not have too much redundant content that is repetitive of the Plan.",

    "Plan": "Based on the [Requirements], [Environmental Information], Thought, context, and user mission, provide a planning scheme.

Note:

1. When involving multiple tool calls, provide the overall plan and the plan for the first action during the first Plan, and provide the plan for the current step in subsequent dialogues.

2. The Plan is a general explanation of the Thought. The Plan does not need to set the values of the tool parameters; it only needs to explain which tools should be called to complete what missions, only the purpose of calling the tools.

3. The format of the Plan needs to be consistent with the example given in the [Requirements].

4. Do not have too much redundant content that is repetitive of the Thought.",

    "Action\_List": [

        \parindent=4em
        
        \{  

            \parindent=6em
            
            "name": "Based on the [Requirements], provide the action to be taken, i.e., the selected tool name",
            
            "arguments": "Based on the [Requirements], [Environmental Information], and [Tool List], provide the input parameters for the action to be taken, i.e., the tool's input parameters. Note: 1. Optional parameters not specified by the user do not need to be provided. 2. Use the JSON format in terms of format, use a dictionary object, do not use strings, and there is no need to provide comments for the parameters",
            
            "tool\_call\_purpose": "The purpose of the tool call"
            
        \parindent=4em
        \}

    \parindent=2em
    ]

\noindent\}

\noindent"""

\end{tcolorbox}
\caption{Planner Decision Generation Prompt Part-2.}
\label{P-planner_2}
\end{figure*}

\begin{figure*}[htbp]

\begin{tcolorbox}[title={Planner Decision Generation Prompt Part-3.}]

[Requirements]="""

*** Special Attention ***

1. When making a decision, please ensure that the tool you invoke from the [Tool List] is suitable for solving the user's mission based on the definition of the tools in the list. Do not force the use of inappropriate tools to solve the user's mission; instead, call the appropriate tool from the [Tool List] according to the user's mission.

\medskip

2. Ensure that the Action\_List you provide does not contradict the Plan you have set out. The order of tools in the given Action\_List should be consistent with the sequence planned in the Plan.

\medskip

3. For optional parameters, you only need to fill them in if the user has provided a value that is different from the default or if there is no default value. Otherwise, there is no need to include them in the arguments.

\medskip

*** The prepare\_to\_answer tool needs to be called in the following two scenarios: ***

1. If you believe that the user's mission can be completed, then call the prepare\_to\_answer tool to provide a summary response, with the answer\_type parameter set to `tool'.

\medskip

2. If you believe that the user's mission does not require the use of any tools from the [Tool List] or that there is no suitable tool to solve the user's mission and it can be answered directly, then call the prepare\_to\_answer tool, with the answer\_type parameter set to `chat'.

\medskip

Note:

1) The absence of a suitable tool in the [Tool List] to solve the user's mission does not mean that you lack the ability to answer. Please respond based on the context information and the knowledge you possess. Do not excessively refuse to answer, nor imagine knowledge you do not have. Only refuse to answer when you cannot respond based on the context information and your own knowledge.

\medskip

2) The absence of a suitable tool in the [Tool List] to solve the user's mission also includes the following situation:

\medskip

First, analyze the common entities that appear in each tool. For example, some tools can only query data related to a certain entity A. If the user asks about entity B, it also means that there is no suitable tool.

\medskip

For instance:

- If the tools in the [Tool List] can only query and analyze population data for Denmark, and the user asks for population data for Sweden, then you should also call the prepare\_to\_answer tool.

\medskip

- If the tools in the [Tool List] can only query weather data for China, including current and historical weather, and the user asks for weather data for the United States, then you should also call the prepare\_to\_answer tool.

\end{tcolorbox}
\caption{Planner Decision Generation Prompt Part-3.}
\label{P-planner_3}
\end{figure*}

\begin{figure*}[htbp]

\begin{tcolorbox}[title={Planner Decision Generation Prompt Part-4.}]

*** There are four scenarios in which the ask\_user\_for\_required\_parameters tool needs to be invoked: ***

1. If you believe that a user's mission requires the use of a tool from the [Tool List], but the user's mission is missing some required parameters from the tool, and the user needs to provide the necessary information, then invoke the ask\_user\_for\_required\_parameters tool. Please do not hallucinate parameters.

\medskip

2. Please note that you are unable to deduce the values of some tool parameters on your own and will need to invoke the ask\_user\_for\_required\_parameters tool to ask the user. Please do not hallucinate parameters.

\medskip

For example:

1) For the timestamp parameter, you do not have the ability to deduce the timestamp based on time. However, you can deduce other time-related parameters (start\_time, end\_time, etc.) on your own based on [Environmental Information], without needing to invoke the ask\_user\_for\_required\_parameters tool.

2) For ID-type parameters (station\_id, product\_id, etc.), you do not have the ability to deduce the corresponding ID based on the name.

\medskip

3. Based on the context of the conversation, if you have already asked the user once to provide the necessary information but the user still has not provided all the required parameters, then please continue to invoke the ask\_user\_for\_required\_parameters tool.

\medskip

4. If the user provides incomplete parameter values, such as the tool parameter being an IP address (ip\_address), but the user provides an incomplete IP address (e.g., 192.22), please continue to use the ask\_user\_for\_required\_parameters tool to ask the user for the complete IP address.

\medskip

Finally, if you confirm the need to invoke the ask\_user\_for\_required\_parameters tool, provide the inquiry plan in the format: "Ask the user to provide xxx, in order to invoke the xxx tool to xxx" in the Plan.

\end{tcolorbox}
\caption{Planner Decision Generation Prompt Part-4.}
\label{P-planner_4}
\end{figure*}

\begin{figure*}[htbp]

\begin{tcolorbox}[title={Planner Decision Generation Prompt Part-5.}]

*** There are eight scenarios in which multiple tools need to be invoked: ***

If a user mission involves invoking multiple tools, please first analyze the dependency relationships between the multiple invocation tools. For tools that do not have invocation dependencies, perform concurrent invocations, and for tools that do have invocation dependencies, perform serial invocations.
Specifically, you can handle each of the following eight scenarios separately:

\medskip

Concurrent invocation scenarios:

\medskip

1. If you determine that the user mission requires multiple invocations of the same tool A, but with different parameters for each invocation of tool A, then please invoke tool A concurrently and provide the concurrent invocation plan in the Plan in the format: "Concurrently invoke tool A N times for xxx."

\medskip

2. If you determine that the user mission requires the invocation of different tools, such as tools A and B, and there is no dependency between tool A and B, then please invoke tools A and B concurrently, and provide the concurrent invocation plan in the Plan in the format: "Concurrently invoke tool A for xxx, while invoking tool B for xxx."

\medskip

Serial invocation scenarios:

\medskip

3. If you determine that the user mission requires the invocation of different tools, such as tools A, B, and C, and there are dependencies between these tools, then please invoke tools A, B, and C serially, and provide the serial invocation plan in the Plan in the format: "First, invoke tool A for xxx. Then, invoke tool B for xxx. Next, invoke tool C for xxx. Now, I will invoke tool A for xxx."

\medskip

Serial invocation has the following two dependency scenarios:

\medskip

3.1. Parameter dependency: For example, before invoking tool C, it is necessary to first invoke tool B to obtain the result as an input parameter, and before invoking tool B, it is necessary to first invoke tool A to obtain the result as an input parameter. Therefore, you need to first complete the invocation of tool A to obtain the result, use it as the input parameter for invoking tool B, and after obtaining the result from tool B, use it as the input parameter for invoking tool C, i.e., please invoke tools A, B, and C serially.

\medskip

3.2. Logical dependency: Even if there is no parameter dependency between the invocation of tools A, B, and C, but there is a logical dependency, such as logically needing to invoke tool B before tool C, and tool A before tool B, then please also invoke tools A, B, and C serially.

\end{tcolorbox}
\caption{Planner Decision Prompt Generation Part-5.}
\label{P-planner_5}
\end{figure*}

\begin{figure*}[htbp]

\begin{tcolorbox}[title={Planner Decision Generation Prompt Part-6.}]

Combined serial and concurrent invocation scenarios:

\medskip

4. If you determine that the user mission requires the invocation of different tools, such as tools A, B, and C, and tool C depends on the invocation of tools A and B, but there is no dependency between tools A and B, then please invoke tools A and B concurrently, followed by the serial invocation of tool C, and provide the combined serial and concurrent invocation plan in the Plan in the format: "Concurrently invoke tools A and B for xxx and xxx, respectively. Then, invoke tool C for xxx. Now, I will concurrently invoke tools A and B for xxx and xxx."

\medskip

5. If you determine that the user mission requires the invocation of different tools, such as tools A, B, and C, and tools B and C depend on the invocation of tool A, but there is no dependency between tools B and C, then please first invoke tool A serially, followed by the concurrent invocation of tools B and C, and provide the combined serial and concurrent invocation plan in the Plan in the format: "First, invoke tool A for xxx. Then, concurrently invoke tools B and C for xxx and xxx, respectively. Now, I will invoke tool A for xxx."

\medskip

6. If you determine that the user mission requires the invocation of different tools, such as tools A and B, and there is a dependency between tools A and B, and tool A needs to be invoked multiple times, then please first invoke tool A concurrently multiple times, followed by the serial invocation of tool B, and provide the combined serial and concurrent invocation plan in the Plan in the format: "First, concurrently invoke tool A N times for xxx. Then, invoke tool B for xxx. Now, I will concurrently invoke tool A N times for xxx."

\medskip

7. If you determine that the user mission requires the invocation of different tools, such as tools A and B, and there is a dependency between tools A and B, and tool B needs to be invoked multiple times, then please first invoke tool A serially, followed by the concurrent invocation of tool B multiple times, and provide the combined serial and concurrent invocation plan in the Plan in the format: "First, invoke tool A for xxx. Then, concurrently invoke tool B N times for xxx. Now, I will invoke tool A for xxx."

\medskip

Special scenarios:

\medskip

8. The tools prepare\_to\_answer and ask\_user\_for\_required\_parameters cannot be invoked concurrently with other tools and need to be invoked serially.

\end{tcolorbox}
\caption{Planner Decision Generation Prompt Part-6.}
\label{P-planner_6}
\end{figure*}

\begin{figure*}[htbp]

\begin{tcolorbox}[title={Planner Decision Generation Prompt Part-7.}]

Please also note:

\medskip

1. The dependency relationship between tool invocations refers to the necessity of completing the call to Tool A before running the call to Tool B.

\medskip

2. For multiple invocations of the same tool, it is necessary to carefully analyze the dependency relationship of each call, noting that even two calls to the same tool may be interdependent.

\medskip

3. If you state in your Thought and Plan that tools need to be called in sequence, then the number of tools to be called in your given Action\_List cannot exceed one, otherwise, there will be a logical contradiction!

\medskip

4. If you cannot ensure that parallel calls to multiple tools A, B, C will not have parameter dependencies and logical dependencies, then please call multiple tools A, B, C in sequence!

\medskip

*** Special Circumstances ***

\medskip

In the following three cases, there is no need to call the ask\_user\_for\_required\_parameters tool:

\medskip

1. If a tool's parameter is a country's ISO code, and the user's mission mentions a specific country, such as China, you can directly deduce China's ISO code and fill it in.

\medskip

2. If a tool's parameter is a longitude or latitude value, and the user's mission mentions a specific location, such as Beijing, you can directly deduce the approximate longitude and latitude values for Beijing and fill them in.

\medskip

3. If a tool's parameter is a time-related parameter (such as start\_time, end\_time, or other parameters that include year, month, and day) and not a timestamp type, you can deduce it based on the current time in the [Environmental Information] and fill it in. At the same time, you need to explain in your Thought how you deduced the value of the time-related parameter based on the current time.

\medskip

*** Other Notes: ***

\medskip

1. Be sure not to provide comments for parameters, as providing parameter comments will cause JSON to fail to parse.

"""

\medskip

\{\{\{all\_tool\_required\_info\}\}\}

\medskip

[Tool List]="""

\{\{\{tools\}\}\}

"""

\end{tcolorbox}
\caption{Planner Decision Generation Prompt Part-7.}
\label{P-planner_7}
\end{figure*}

\begin{figure*}[htbp]

\begin{tcolorbox}[title={Tool Feedback Generation Prompt.}]

Please act as an external tool, Tool, within a super intelligent agent. These external tools can be used to solve user missions, as detailed in the [Tool List].

\medskip

Based on the tool name and input parameters output by the super intelligent agent's Planner, simulate the execution results of the tool.

\medskip

If there are multiple tools in the Action\_List provided by the Planner, please simulate each one separately, ensuring the number matches the Action\_List, and store the results in the Observation\_List.
Refer to the [Tool Output Format] for the output format.

\medskip

[Environmental Information]="""

\{\{\{env\_info\}\}\}

"""

\medskip

\medskip

[Tool Invocation Result Requirements]="""

1. Validate the HTTP method and parameters in the request according to the OpenAPI specification.

2. Generate a response that strictly follows the format specified in the OpenAPI specification and ensure it is in JSON format.

3. The response should contain real data, avoiding the use of placeholders.

4. Handle edge cases by providing appropriate error responses.

5. For requests without length limitations, such as the GET method, ensure the response returns 3 to 5 samples, and be careful not to use ellipses like // xxx, ... to omit sample information, as it must conform to JSON format, otherwise it will cause JSON parsing errors!

6. Try to simulate responses in English.

"""

\medskip

[Tool List]="""

\{\{\{tools\}\}\}

"""

\medskip

[Tool Output Format]="""

Tool:

\{

    \parindent=2em

    "Observation\_List": [

        \parindent=4em
        
        \{

            \parindent=6em
        
            "status\_code": "Refer to [Tool Invocation Result Requirements] for the HTTP response status code",
            
            "response": "Refer to [Tool Invocation Result Requirements] to simulate the result of the action execution. Ensure your response is in JSON format, contains real data, and complies with the OpenAPI specification format."

        \parindent=4em    
        \}

    \parindent=2em
    ]
 
\noindent
\}

\noindent
"""

\end{tcolorbox}
\caption{Tool Feedback Generation Prompt.}
\label{P-tool}
\end{figure*}

\begin{figure*}[htbp]
\begin{tcolorbox}[title={AI Feedback Generation Prompt.}]

Please act as an Agent assistant within a super intelligent agent, which has a series of external tools. The Planner within the super intelligent agent can solve user missions by calling external tools, as detailed in the [Tool List].

\medskip

You are responsible for interacting with the user. Based on the results returned by the Planner and Tool, combined with the user mission and the context of the conversation, you provide answers, and only your answers will be displayed to the user.

\medskip

Refer to the [Agent Assistant Output Format] for the output format.

\medskip

[Environmental Information]="""

\{\{\{env\_info\}\}\}

"""

\medskip

[Agent Assistant Output Format]="""

Agent Assistant: According to the [Requirements], reply to the most recent round of content starting with "User:" in the context conversation information (do not repeat this sentence).

"""

\medskip

[Requirements]="""

1. The reply must start with "Agent Assistant:".

2. Summarize the user mission from the most recent round starting with "User:" based on the context conversation information.

3. Use markdown format, and be sure to pay attention to the layout to make it look neat, with two line breaks between paragraphs.

4. Pay special attention! If the Observation given by the Tool is a list, and each item in the list has its own ID, such as xxx\_id or xxxId, then when summarizing the reply, please retain these IDs for each item and inform the user!

5. Reply in English.

"""

\medskip

\{\{\{all\_tool\_required\_info\}\}\}

\medskip

[Tool List]="""

\{\{\{tools\}\}\}

"""

\end{tcolorbox}
\caption{AI Feedback Generation Prompt.}
\label{P-AI}

\end{figure*}

\end{document}